\documentclass[sn-nature,iicol]{sn-jnl}

\usepackage{graphicx}%
\usepackage{multirow}%
\usepackage{amsmath,amssymb,amsfonts}%
\usepackage{amsthm}%
\usepackage{mathrsfs}%
\usepackage[title]{appendix}%
\usepackage{xcolor}%
\usepackage{textcomp}%
\usepackage{manyfoot}%
\usepackage{booktabs}%
\usepackage{algorithm}%
\usepackage{algorithmicx}%
\usepackage{algpseudocode}%
\usepackage{listings}%
\usepackage[version=4]{mhchem}
\usepackage{microtype}

\theoremstyle{thmstyleone}%
\theoremstyle{thmstyletwo}%

\theoremstyle{thmstylethree}%

\usepackage{acro}
\newcommand{\acrocolor}[1]{\textcolor{black}{#1}}
\acsetup{make-links=true,format=\acrocolor}

\DeclareAcronym{hsi}{short=HSI,long=hyperspectral imaging}
\DeclareAcronym{ai}{short=AI,long=artificial intelligence}
\DeclareAcronym{ml}{short=ML,long=machine learning}
\DeclareAcronym{iqr}{short=IQR,long=interquartile range}
\DeclareAcronym{dsc}{short=DSC,long=Dice similarity coefficient}
\DeclareAcronym{nsd}{short=NSD,long=normalized surface Dice}
\DeclareAcronym{sto2}{short=\ce{StO2},long=tissue oxygen saturation}
\DeclareAcronym{pca}{short=PCA,long=principal component analysis}
\DeclareAcronym{mitk}{short=\acrocolor{MITK},long=Medical Imaging Interaction Toolkit}
\DeclareAcronym{ce}{short=\acrocolor{CE},long=cross-entropy}
\DeclareAcronym{sd}{short=\acrocolor{SD},long=standard deviation}
\DeclareAcronym{icg}{short=\acrocolor{ICG},long=indocyanine green}

\usepackage{tabularx}
\usepackage{csquotes}
\usepackage{xspace}
\usepackage{siunitx}
\DeclareSIUnit\bpm{bpm}
\DeclareSIUnit\Unit{Unit}
\newcommand{\varTotalImages}{\num{13874}\xspace}
\newcommand{\varTotalSubjects}{\num{319}\xspace}
\newcommand{\varTotalImagesSemantic}{\num{2596}\xspace}
\newcommand{\varTotalImagesPolygon}{\num{11278}\xspace}

\newcommand{\varTotalImagesPhysPig}{\num{3890}\xspace}
\newcommand{\varTotalImagesMalPig}{\num{446}\xspace}
\newcommand{\varTotalImagesICGPig}{\num{556}\xspace}

\newcommand{\varTotalImagesPhysRat}{\num{4681}\xspace}
\newcommand{\varTotalImagesMalRat}{\num{1538}\xspace}
\newcommand{\varTotalImagesICGRat}{\num{1137}\xspace}

\newcommand{\varTotalSubjectsHuman}{\num{230}\xspace}
\newcommand{\varTotalImagesPhysHuman}{\num{830}\xspace}
\newcommand{\varTotalImagesMalHuman}{\num{796}\xspace}

\newcommand{\varDSCInSpeciesPig}{0.90 (\ac{sd} 0.09)\xspace}
\newcommand{\varDSCInSpeciesRat}{0.90 (\ac{sd} 0.09)\xspace}
\newcommand{\varDSCInSpeciesHuman}{0.78 (\ac{sd} 0.17)\xspace}
\newcommand{\varHSIDropRanging}{\SI{-44}{\percent} (pig2rat) to \SI{-56}{\percent} (rat2human)\xspace}
\newcommand{\varDistanceImprovementRatToPig}{\SI{29.7}{\percent} (\ac{sd} \SI{ 12}{\percent})\xspace}
\newcommand{\varDistanceImprovementPigToRat}{\SI{27.6}{\percent} (\ac{sd} \SI{ 12}{\percent})\xspace}
\newcommand{\varDistanceImprovementRatToHuman}{\SI{26.5}{\percent} (\ac{sd} \SI{ 8}{\percent})\xspace}
\newcommand{\varDistanceImprovementPigToHuman}{\SI{20.5}{\percent} (\ac{sd} \SI{ 6}{\percent})\xspace}
\newcommand{\varICGImprovementMeanPig}{0.39 (\ac{sd} 0.28)\xspace}
\newcommand{\varICGImprovementMaxPig}{0.96\xspace}
\newcommand{\varICGImprovementMaxLabelPig}{liver\xspace}
\newcommand{\varICGImprovementMeanRat}{0.28 (\ac{sd} 0.19)\xspace}
\newcommand{\varICGImprovementMaxRat}{0.88\xspace}
\newcommand{\varICGImprovementMaxLabelRat}{liver\xspace}

\newcommand{\autocite}[1]{\citep{#1}}

\begin{document}

\title[Xeno-learning: knowledge transfer across species in deep learning-based spectral image analysis]{Xeno-learning: knowledge transfer across species in deep learning-based spectral image analysis}


\author[1,2,4]{\fnm{Jan} \sur{Sellner}}
\equalcont{These authors contributed equally to this work.}
\author[5,6,7,8]{\fnm{Alexander} \sur{Studier‑Fischer}}
\equalcont{These authors contributed equally to this work.}
\author[1,2,3]{\fnm{Ahmad} \spfx{Bin} \sur{Qasim}}
\author[1,2,3,4]{\fnm{Silvia} \sur{Seidlitz}}
\author[9]{\fnm{Nicholas} \sur{Schreck}}
\author[1]{\fnm{Minu} \sur{Tizabi}}
\author[9]{\fnm{Manuel} \sur{Wiesenfarth}}
\author[9]{\fnm{Annette} \sur{Kopp‑Schneider}}
\author[1,11]{\fnm{Janne} \sur{Heinecke}}
\author[1,11]{\fnm{Jule} \sur{Brandt}}
\author[5]{\fnm{Samuel} \sur{Knödler}}
\author[6,7,8]{\fnm{Caelan} \spfx{Max} \sur{Haney}}
\author[5]{\fnm{Gabriel} \sur{Salg}}
\author[5,7,8]{\fnm{Berkin} \sur{Özdemir}}
\author[10]{\fnm{Maximilian} \sur{Dietrich}}
\author[6,7,8]{\fnm{Maurice} \spfx{Stephan} \sur{Michel}}
\author[12]{\fnm{Felix} \sur{Nickel}}
\author[6,7,8]{\fnm{Karl-Friedrich} \sur{Kowalewski}}
\author*[1,2,3,4,11]{\fnm{Lena} \sur{Maier‑Hein}}\email{l.maier-hein@dkfz-heidelberg.de}

\affil*[1]{Division of Intelligent Medical Systems (IMSY), German Cancer Research Center (DKFZ), Heidelberg, Germany}
\affil[2]{Helmholtz Information and Data Science School for Health, Heidelberg/Karlsruhe, Germany}
\affil[3]{Faculty of Mathematics and Computer Science, Heidelberg University, Heidelberg, Germany}
\affil[4]{National Center for Tumor Diseases (NCT), NCT Heidelberg, a partnership between DKFZ and university medical center Heidelberg}
\affil[5]{Department of General, Visceral, and Transplantation Surgery, Heidelberg University Hospital, Heidelberg, Germany}
\affil[6]{Department of Urology and Urosurgery, Medical Faculty of the University of Heidelberg, University Medical Center Mannheim, Mannheim, Germany}
\affil[7]{Division of Intelligent Systems and Robotics in Urology (ISRU), German Cancer Research Center (DKFZ) Heidelberg, Heidelberg, Germany}
\affil[8]{DKFZ Hector Cancer Institute at the University Medical Center Mannheim, Mannheim, Germany}
\affil[9]{Division of Biostatistics, German Cancer Research Center (DKFZ), Heidelberg, Germany}
\affil[10]{Heidelberg University, Medical Faculty, Department of Anesthesiology, Heidelberg University Hospital, Heidelberg, Germany}
\affil[11]{Medical Faculty, Heidelberg University, Heidelberg, Germany}
\affil[12]{Department of General, Visceral and Thoracic Surgery, University Hospital Hamburg-Eppendorf, Germany}

\abstract{Novel optical imaging techniques, such as \acf*{hsi} combined with machine learning-based (ML) analysis, have the potential to revolutionize clinical surgical imaging. However, these novel modalities face a shortage of large-scale, representative clinical data for training ML algorithms, while preclinical animal data is abundantly available through standardized experiments and allows for controlled induction of pathological tissue states, which is not ethically possible in patients. To leverage this situation, we propose a novel concept called \enquote{xeno-learning}, a cross-species knowledge transfer paradigm inspired by xeno-transplantation, where organs from a donor species are transplanted into a recipient species. Using a total of \varTotalImages \acs*{hsi} images from humans as well as porcine and rat models, we show that although spectral signatures of organs differ substantially across species, relative changes resulting from pathologies or surgical manipulation (e.g., malperfusion; injection of contrast agent) are comparable. Such changes learnt in one species can thus be transferred to a new species via a novel \enquote{physiology-based data augmentation} method, enabling the large-scale secondary use of preclinical animal data for humans. The resulting ethical, monetary, and performance benefits promise a high impact of the proposed knowledge transfer paradigm on future developments in the field.}


\maketitle
\section{Main}
Death within 30 days after surgery has been found to be the third-leading contributor to mortality worldwide \autocite{Nepogodieva2019}. One of the major challenges faced by surgeons is the visual discrimination of tissues, for example to distinguish between pathological and physiological tissue or spare critical intraoperative structures. Spectral imaging has been proposed as a means of overcoming the limitations of visual perception \autocite{10.1016/j.media.2020.101699}. While conventional medical cameras (e.g., laparoscopic imaging systems) are limited by \enquote{imitating} the human eye and recording only red, green and blue colors, spectral cameras remove this arbitrary restriction and instead capture multiple specific bands of light that allow for decoding relevant information on tissue type and function \autocite{10.1038/s41598-022-15040-w,10.1515/bmt-2017-0155}.

\Ac{ai} in general and \ac{ml} in particular have evolved as key enabling techniques to convert the high-dimensional spectral data into clinically useful information \autocite{10.3390/technologies12090163,10.1109/ACCESS.2021.3068392,10.1016/j.pdpdt.2020.102165,10.3390/s21186002}. However, a major bottleneck in converting this novel imaging technique’s potential into patient benefit lies in the lack of large annotated, high-quality datasets covering the wide range of pathologies that can occur in practice. In this context, preclinical animal data represents an untapped resource, offering not only more data but also the possibility of conducting various types of standardized experiments. This is in stark contrast to human data acquisition for which high ethical standards and clinical workflow considerations render many experiments of high scientific interest infeasible, such as the intentional induction of pathological tissue states, e.g., fibrosis, inflammation or malperfusion.

\begin{figure*}[ht]
    \centering
    \includegraphics[width=0.9\linewidth]{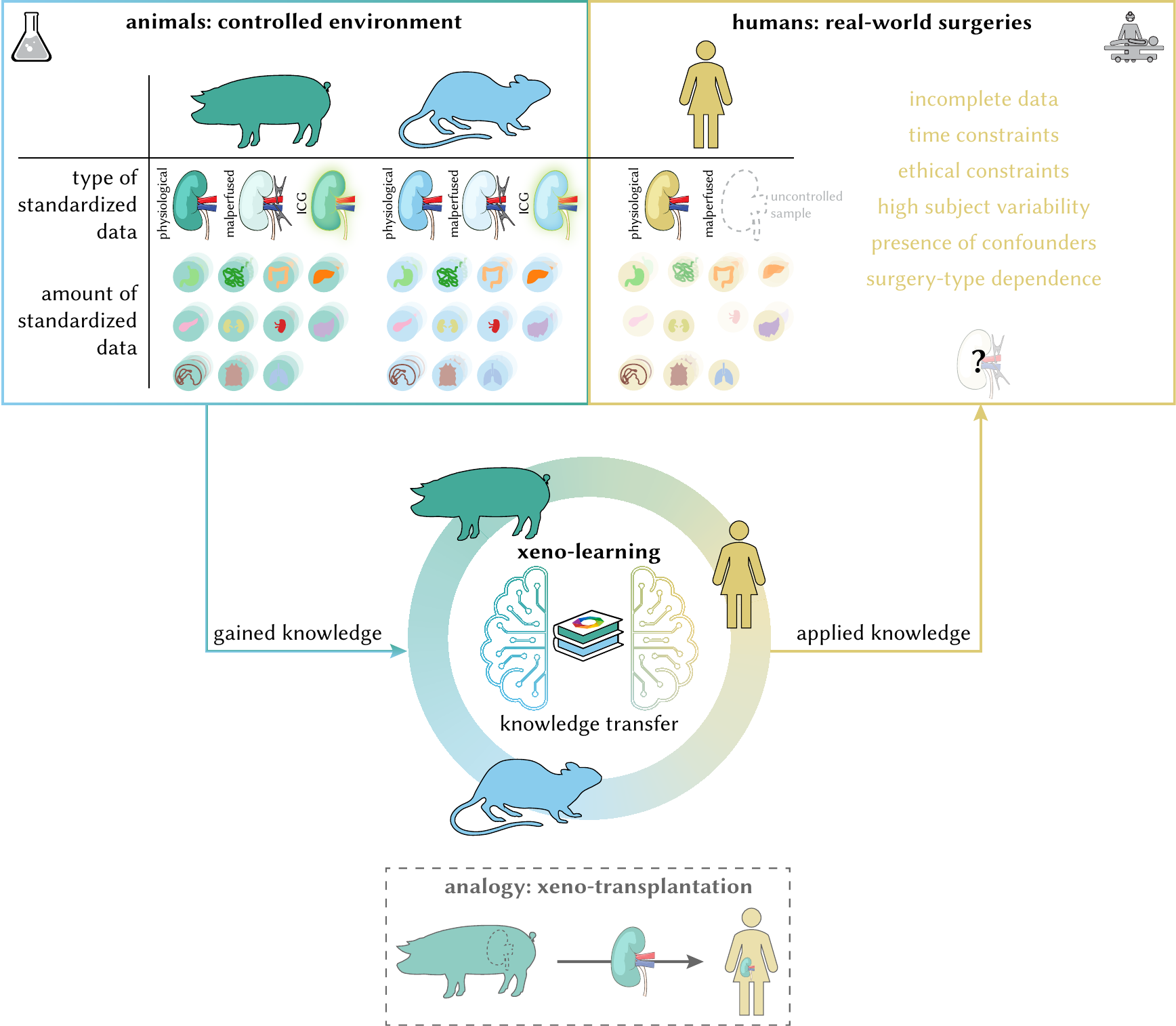}
    \caption{\textbf{Inspired by the concept of xeno-transplantation, i.e., transplanting an organ from one species to another, we propose xeno-learning as a new paradigm for transferring knowledge across species.} Human data obtained with novel imaging modalities, such as \acf*{hsi}, is often sparse, does not represent the broad range of pathologies or surgical manipulations (e.g., malperfusion; injection of contrast agent) that can occur in practice and lacks standardization due to ethical and clinical workflow issues. We introduce the method of xeno-learning as a new paradigm to boost the performance of neural networks applied to one species by making efficient use of data from another species.}
    \label{fig:intro}
\end{figure*}

\begin{figure*}[htp]
    \centering
    \includegraphics[width=0.9\linewidth]{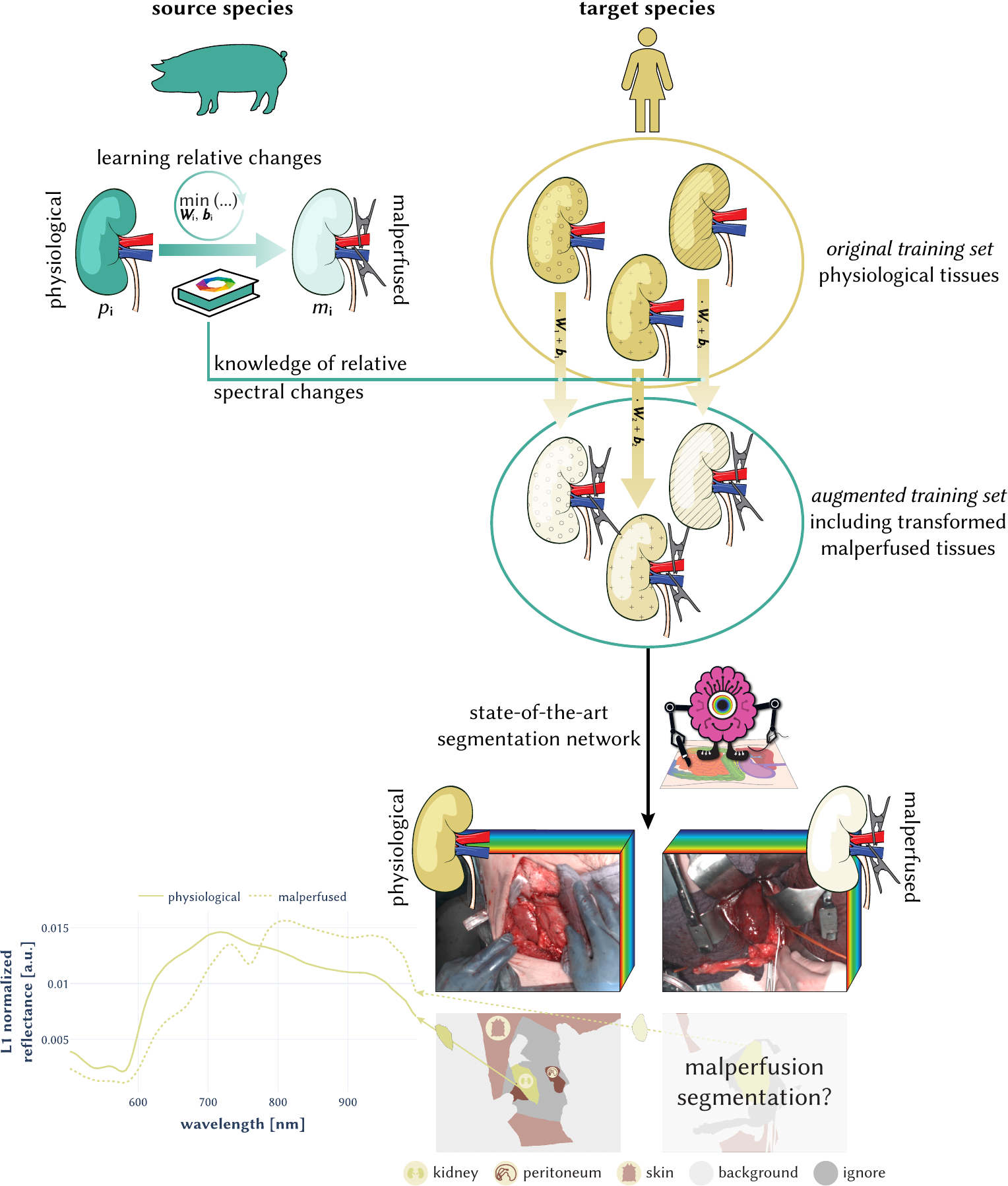}
    \caption{\textbf{Knowledge on perfusion shifts is transferred via a novel physiology-based data augmentation method.} The example shows how relative spectral changes resulting from perfusion shifts are learnt from physiological and malperfused tissue pairs $(p_i, m_i)$ of the source species and encoded in a linear model. The target species’ training set, which contains only physiological tissues, is augmented by applying the learnt model to the target spectra. State-of-the-art segmentation networks \autocite{10.1016/j.media.2022.102488,10.1007/978-3-031-43996-4_59,10.48550/arxiv.2408.15373} are applied for surgical scene segmentation. The example images show human physiological and malperfused kidneys, their corresponding semantic segmentation and median spectra for the annotated kidney regions. The example focuses on pig as the source and human as the target species, with malperfusion as data shift. However, the same method can also be applied to other species or data shifts such as  those arising from the application of contrast agents, e.g., \acf*{icg}.}
    \label{fig:method_concept}
\end{figure*}

To address the human data bottleneck, we therefore propose the new concept of \enquote{xeno-learning} (cf.\ \autoref{fig:intro}), a new \ac{ai} paradigm to transfer knowledge across species. The term has been inspired by the concept of xeno-transplantation, which refers to the transplantation of living cells, tissues, or organs from one species to another. In this case, the shortage of transplantable human organs is addressed by using organs from other species; in analogy, our work aims to address the shortage of standardized data by transferring knowledge from one species (specifically porcine or rat models) to another (here: humans). Our specific contributions are:

\begin{itemize}
    \item \emph{Knowledge transfer bottleneck}: We demonstrate that spectral signatures of organs differ across species, thus causing neural networks trained on one species (e.g., animals) to fail when classifying tissues of another species (e.g., human tissue).
    \item \emph{New cross-species learning paradigm}: We introduce the concept of xeno-learning as a new paradigm to boost the performance of neural networks applied to one species by making efficient use of data from another species. We further instantiate the concept on the specific challenge of perfusion shifts in spectral image analysis, which may lead to radical performance decrease in state-of-the-art tissue segmentation methods. Relative spectral changes are learnt in one species and transferred to a new species via a novel \enquote{physiology-based data augmentation} method (cf.\ \autoref{fig:method_concept}).
    \item \emph{Comprehensive validation with three species}: Based on an \ac{hsi} database of unprecedented size, comprising \varTotalImages hyperspectral images from humans as well as porcine and rat models, we show that tissue discrimination performance can be boosted by learning from another species.
\end{itemize}

\section{Results}
Both medicine in general and surgery in particular offer numerous examples where animal research has significantly informed clinical practice \autocite{10.1177/0023677216642398}. The success of animal experiments can largely be attributed to existing similarities between humans and certain species, such as rats and pigs. In the following, we explore to which extent spectral images acquired from animals can be leveraged for human applications.

All of the presented results were obtained using the medical device-graded \ac{hsi} system Tivita\textsuperscript{\textregistered} Surgery (Diaspective Vision, Am Salzhaff, Germany) to collect \varTotalImages \ac{hsi} images from three species including \varTotalSubjectsHuman patients from Heidelberg University Hospital. For our analysis, we semantically annotated \varTotalImagesSemantic images with 12 classes (11 organ classes and background).

\subsection{Differences in organ spectra result in segmentation network failure to generalize across species}
\begin{figure*}[ht]
    \centering
    \includegraphics[width=\linewidth]{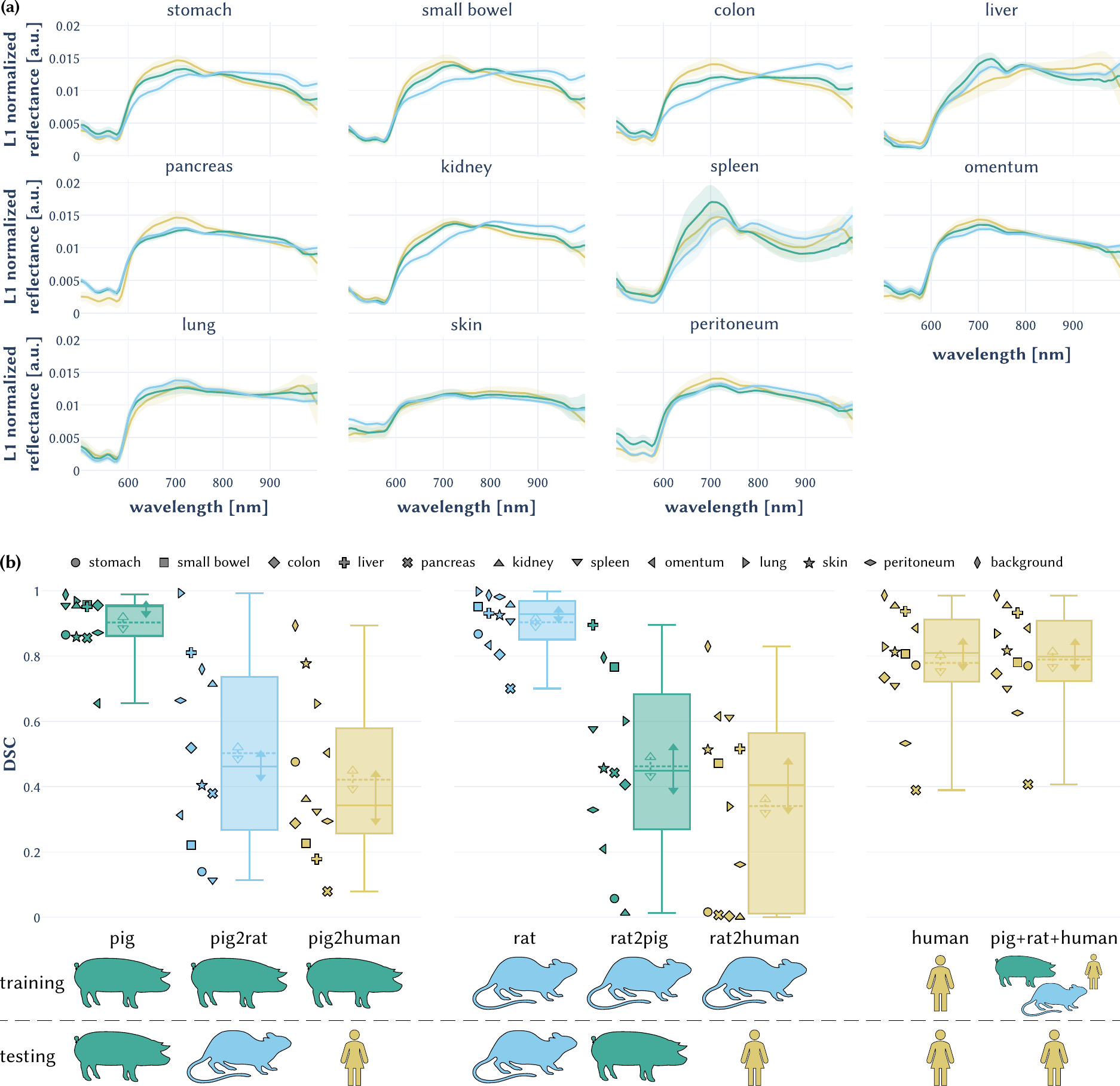}
    \caption{\textbf{Different species exhibit different spectral organ fingerprints, resulting in organ segmentation networks trained on one species failing to generalize towards unseen species.} \textbf{(a)} For each organ and species, the median spectrum (solid line) and the standard deviation (shaded area) across subjects is shown. The median spectrum was calculated per channel for each annotated region in all images. \textbf{(b)} For each species, the intra-species segmentation performance always surpasses the out-of-species performance. The distributions of hierarchically aggregated class-level \acf*{dsc} scores are shown. Each boxplot depicts the \acf*{iqr} with the median (solid line) and mean (dotted line). Arrows indicate the \SI{95}{\percent} confidence interval of the median and mean based on bootstrapped subject-sampling. The whiskers extend up to 1.5 times of the \acs*{iqr}. Results for the \acf*{nsd} are presented in \autoref{fig:domain_shift_performance_nsd}.}
    \label{fig:domain_shift_performance_dsc}
\end{figure*}

While similarities do exist, spectral organ fingerprints were generally not consistent across species. \autoref{fig:domain_shift_performance_dsc} (a) shows spectral organ fingerprints for a total of 11 organs for three species, namely humans, pigs and rats. Some organs such as skin or omentum exhibited higher similarities in their spectra across species than other organs, such as colon or kidney.

To investigate the effect of the spectral differences on neural network performance, we trained networks on physiological organ images of one species and evaluated their performance on physiological organ images of other species (\autoref{fig:domain_shift_performance_dsc} (b)).

Segmentation networks failed dramatically when applied to a new species, with performance decreases of \varHSIDropRanging. The metric values also varied highly between different classes: Whereas classes such as background or skin could still be detected correctly in many cases, other classes such as pancreas exhibited high performance drops. Joint training of animal and human data (\autoref{fig:domain_shift_performance_dsc} (b): pig+rat+human network) did not improve the discrimination of human tissues.

It should be noted that the intra-species performance values (\ac{dsc}: \autoref{fig:domain_shift_performance_dsc} (b); \ac{nsd} \autoref{fig:domain_shift_performance_nsd}) are consistent with the state-of-the-art literature \autocite{10.1016/j.media.2022.102488,10.1007/978-3-031-43996-4_59,10.48550/arxiv.2408.15373}. Overall, the segmentation performance on human data was lower (\varDSCInSpeciesHuman) than on animal data (\varDSCInSpeciesPig and \varDSCInSpeciesRat for pig and rat, respectively).

\subsection{Extended data analysis provides reasons for neural network failure}
To quantify the relevance of the species relative to other sources of variation, such as the subject or the specific imaging conditions, we performed a mixed effect model analysis. For both pigs and rats, we followed the standardized recording protocol defined in \autocite{10.1038/s41597-023-02315-8}, according to which spectral images from multiple subjects are acquired for a predefined set of recording conditions. For each organ, the proportion of variation in observed reflectance was decomposed into (explained) variation by the factors species, subject, image, and angle, as well as residual or unexplained variation, as shown in \autoref{fig:mixed_effect_analysis}.

To facilitate knowledge transfer, a negligible species effect would be desired. However, this is not the case, with no consistent behavior to be observed for all organs. For several organs, the species even constituted the main source of variability for certain wavelengths. Across all organs, the angle factor and unexplained variation (residuals) only played a minor role.

\begin{figure*}[htp]
    \centering
    \includegraphics[width=\linewidth]{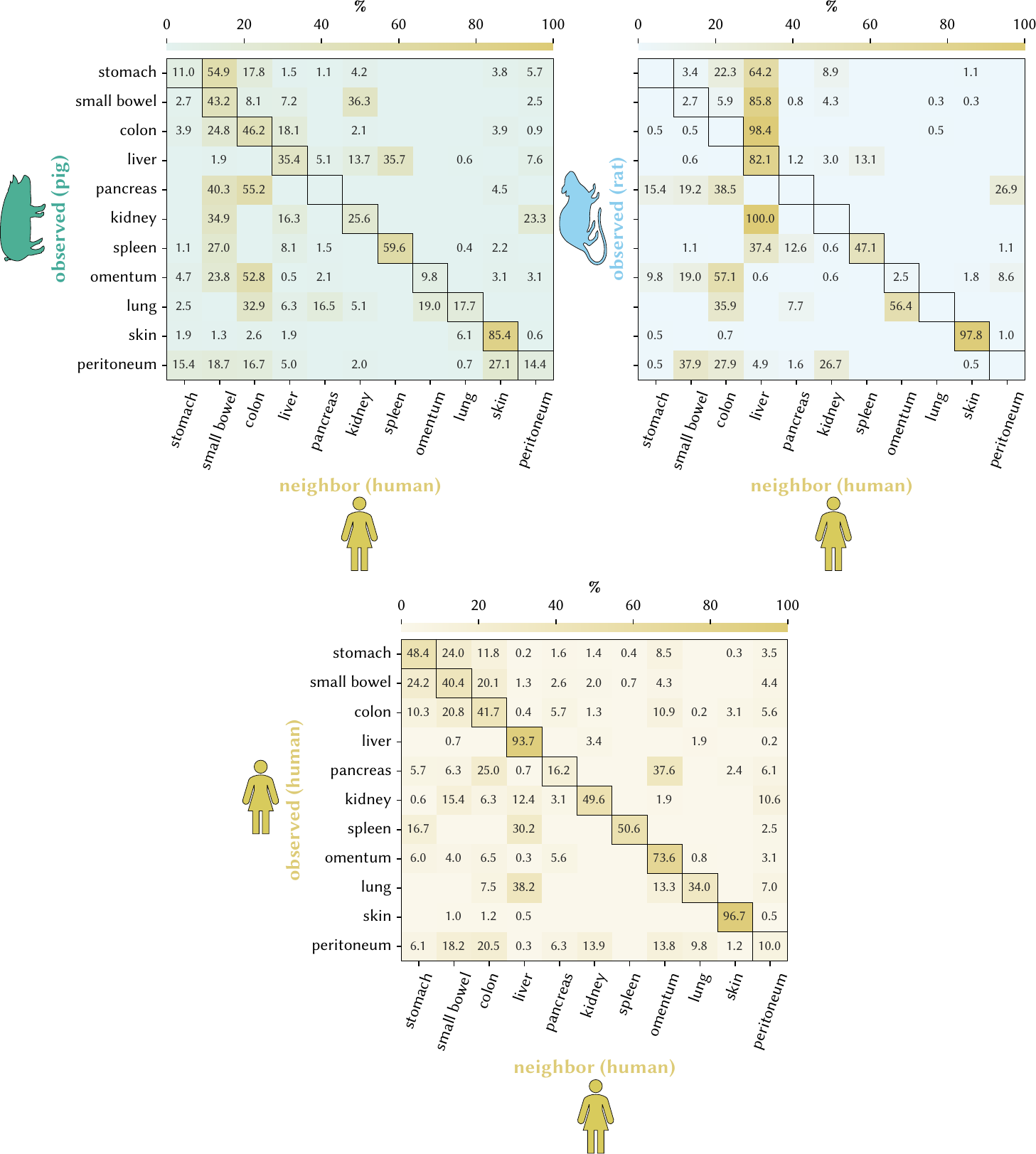}
    \caption{\textbf{Spectra of the same tissue type are not nearest neighbors across species.} Comparison between pig and human (top left), rat and human (top right) and human to human subjects (bottom left) median spectra per organ. For every median spectrum of the observed species, we determined the nearest neighbor in the human species and compared the class labels. The nearest neighbor spectrum is always from a different subject. Each matrix is row-normalized, highlighting how the median spectra from an animal class are distributed across the nearest neighbor human classes. Animal comparisons can be found in \autoref{fig:nearest_neighbor_suppl}.}
    \label{fig:nearest_neighbor}
\end{figure*}

Besides the variability analysis of the spectra, we performed an analysis to examine the extent to which spectral values matched between organs. \autoref{fig:nearest_neighbor} accordingly examines the high-dimensional neighborhood of the median spectra. When comparing organ spectra of one species to those of another, it could be observed that the nearest neighbor only rarely agreed in the organ class (low agreement on the diagonal). For example, porcine spectra of the pancreas were closer to human colon or small bowel spectra than human pancreas spectra. Noticeable exceptions were liver spectra and skin spectra, which were relatively close to each other for all species. In contrast, the nearest neighbors between different human subjects showed a much higher agreement.

\subsection{Shared mechanisms lead to comparable relative spectral changes across species}
Our previous analyses clearly showed that the substantial \emph{absolute differences} of spectral organ signatures across species render naive approaches of knowledge transfer infeasible. Consequently, we investigated whether relative changes in spectra resulting from pathologies (e.g., ischemia) or surgical manipulation (e.g., injection of contrast agent) exhibited similarities across species. We focused on tissue malperfusion, an essential aspect of real-world surgeries, as well as spectral changes following intravenous \ac{icg} injection, an important mechanism for clinical, intraoperative perfusion monitoring. These changes may result in poor tissue discrimination performance for neural networks trained primarily on physiological tissue (cf.\ \autoref{fig:perfusion_performance}). The underlying hypothesis was that the comparable tissue alterations result in similar spectral changes in humans and animals.

\begin{figure*}[htp]
    \centering
    \includegraphics[width=\linewidth]{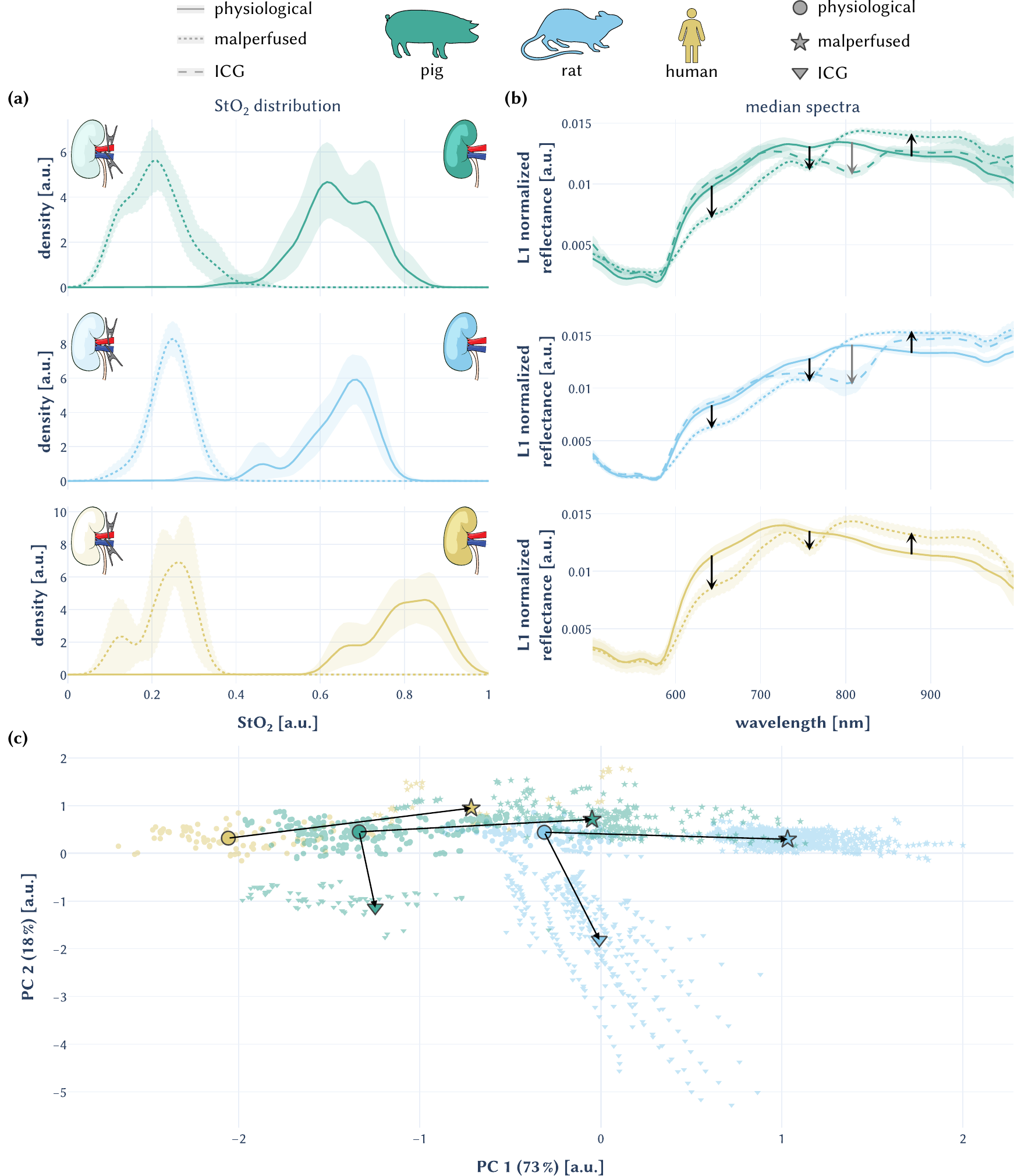}
    \caption{\textbf{Although spectra are different across species, pathologies or surgical manipulations lead to comparable systematic spectral shifts.} For each species, \textbf{(a)} visualizes the distribution of \acf*{sto2} for physiological and malperfused kidney tissues. The corresponding median spectra are illustrated in \textbf{(b)}, highlighting transitions from physiological to malperfused states, as well as shifts in spectra following \acf*{icg} injection. The shaded area highlights the \SI{95}{\percent} confidence interval based on bootstrapped subject-sampling (\acs*{sto2} distribution) and the standard deviation across subjects (median spectra). \textbf{(c)} \Acf*{pca} of the median spectra indicates a spectral shift in the same direction across species when transitioning from physiological to malperfused tissues (in this case, to the right) or from physiological to \acs*{icg}-injected spectra (in this case, to the bottom). Solid markers denote the cluster centers and each transparent marker represents one image whereas the axis labels denote the explained variance of the corresponding principal component.}
    \label{fig:malperfused}
\end{figure*}

We thus systematically compared physiological and malperfused organ spectra as well as spectra following \ac{icg} injection (\autoref{fig:malperfused}). To this end, we focused on kidneys as they provide a highly standardized, biologically unambiguous and clinically relevant model. Although the spectra differed, the change from physiological to malperfused spectra or the changes from physiological to spectra following \ac{icg} injection exhibited a similar pattern across all species. For example, a drop in normalized reflectance could consistently be observed around \SI{750}{\nm} in the case of malperfusion. Low-dimensional projections of the kidney median spectra further illustrate that despite the only minor overlap of physiological samples between species, a similar shift can be observed when transitioning from physiological to malperfused tissue or from physiological to \ac{icg}-injected tissue in \ac{pca} space.

\subsection{Xeno-learning enables knowledge transfer}
Our approach to knowledge transfer is based on the crucial insight that shared mechanisms manifest in comparable relative spectral changes. As illustrated in \autoref{fig:method_concept}, we leverage this insight through data augmentation;  we first learn the effect of a pathology or interventions from data of the source species and then apply this knowledge to the target species. In analogy to physics-based data augmentation, we termed  this method \emph{physiology-based data augmentation}.

According to experimental results obtained on independent test sets that were not used during method development, our proposed xeno-learning method is able to transfer relevant perfusion-related knowledge encoded in preclinical animal data to humans.

\begin{figure*}[htp]
    \centering
    \includegraphics[width=0.98\linewidth]{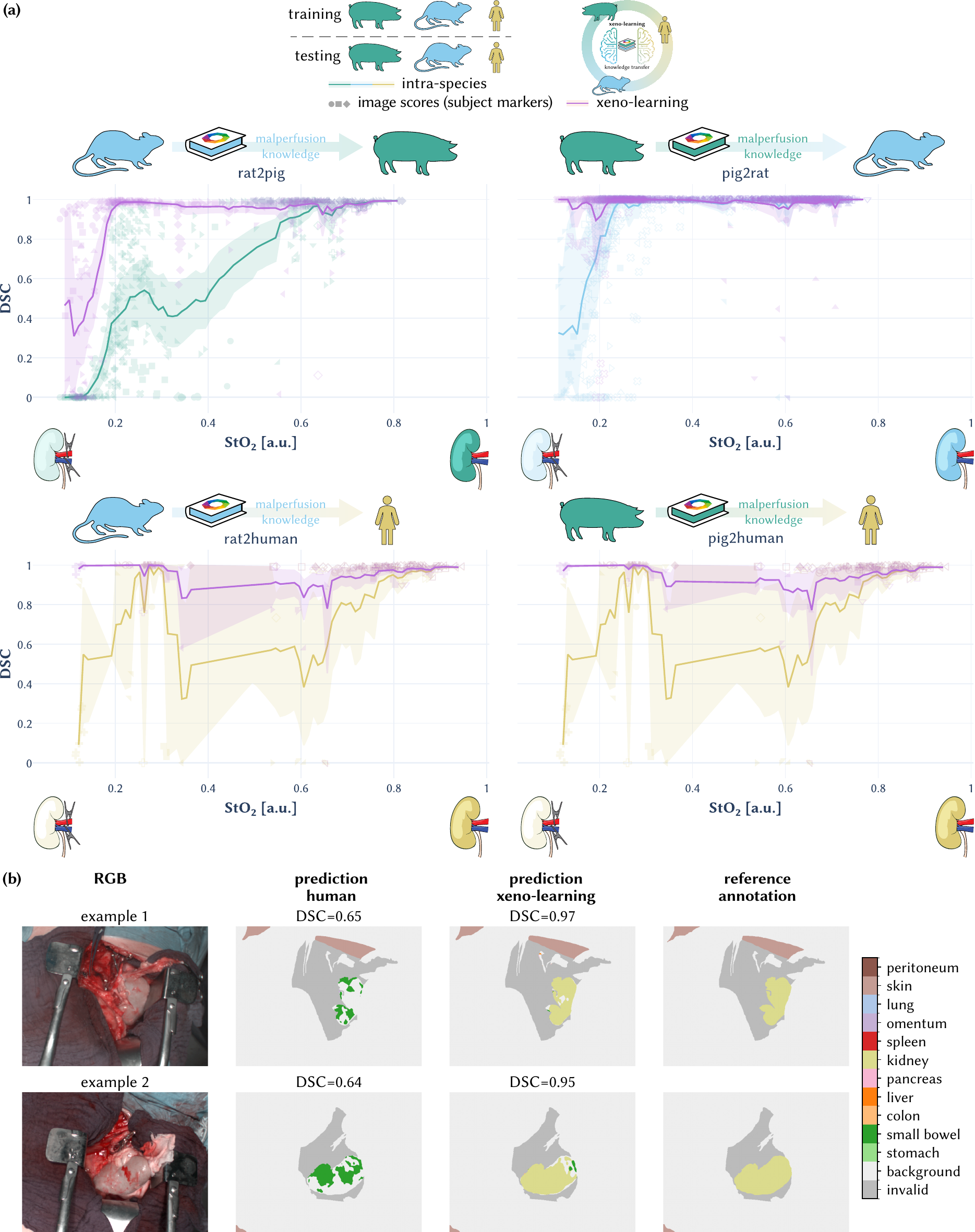}
    \caption{\textbf{Xeno-learning enables knowledge transfer across species.} \emph{(a)} The segmentation performance, measured via the \acf*{dsc}, is substantially improved with the xeno-learning approach, especially for malperfused tissue corresponding to low \acf*{sto2}. For each subject in the target species, a line was interpolated over all $(\acs*{sto2}, \acs*{dsc})$ values and hierarchically aggregated across subjects. The shaded area denotes the \SI{95}{\percent} confidence interval based on bootstrapped subject-sampling and the transparent markers denote the raw kidney scores for individual images (different markers correspond to different subjects). \emph{(b)} The human baseline network fails to correctly classify the kidneys in the example images while the pig2human network, which includes our xeno-learning transformation, segments the kidneys with high accuracy.}
    \label{fig:perfusion_performance}
\end{figure*}

\begin{figure*}[htp]
    \centering
    \includegraphics[width=\linewidth]{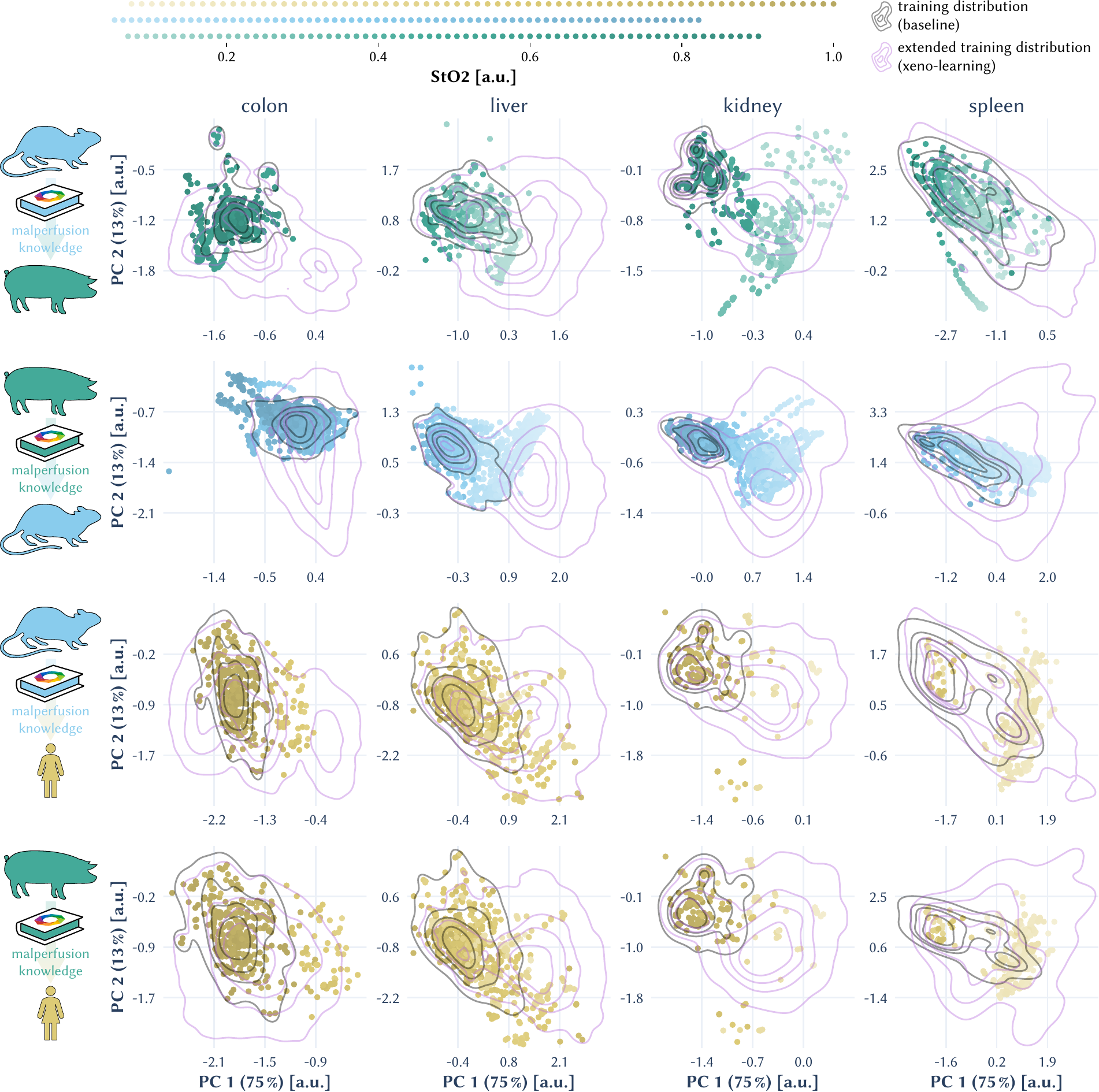}
    \caption{\textbf{Xeno-learning extrapolates the training data to cover a wide range of spectral perfusion states for multiple organs.} \Acf*{pca} of all median spectra from the target species are shown as circles with the lightness encoding the \acf*{sto2} of the tissue. The training distribution of the baseline as well as the extended training distribution obtained via our physiologically-based augmentation was estimated by kernel density estimations and is depicted by gray (original) and purple (xeno-learning) lines. For many organs and species, malperfused spectral data is out-of-distribution with respect to the original training distribution. Xeno-learning resolves this issue.}
    \label{fig:perfusion_projections}
\end{figure*}

As illustrated in \autoref{fig:perfusion_projections}, the original training data (distribution depicted in gray) does not adequately cover the malperfused data (green, blue and yellow points depending on species). Xeno-learning (distribution coloured in purple) resolved the issue. To quantify the effect of our augmentation on the training distribution, we computed the distances between each point in the test dataset to the nearest point in the training dataset. We observed that the distances (averaged over all organs) decreased by \varDistanceImprovementRatToPig for the rat2pig, \varDistanceImprovementPigToRat for the pig2rat, \varDistanceImprovementRatToHuman for the rat2human and \varDistanceImprovementPigToHuman for the pig2human scenarios, when evaluated against the extended training distribution as opposed to the baseline training distribution.

Whether or not a change in spectra resulting from pathophysiological effects results in a drop in the performance of the downstream tasks depends crucially on the specific application. In the case of semantic segmentation the only major drop in performance was observed for the kidney
(\autoref{fig:perfusion_performance}). In general, malperfused tissues present a challenge to the segmentation network across all species and the \ac{dsc} tends to decrease with lower tissue \ac{sto2}. In the case of humans, the knowledge transfer was more beneficial than in rats.

\subsection{Xeno-learning generalizes across knowledge transfer tasks}
To demonstrate that xeno-learning generalizes over knowledge transfer tasks, we prospectively applied our approach to a more complex pathophysiological mechanism without making any changes to the method. Specifically, we investigated whether spectral changes resulting from the injection of the contrast agent \ac{icg} can be learnt across species. In analogy to our first knowledge transfer task,  we encoded relative spectral changes from data without contrast agent to data with \ac{icg} in and transferred them to another species via  data augmentation.

\begin{figure*}[htp]
    \centering
    \includegraphics[width=\linewidth]{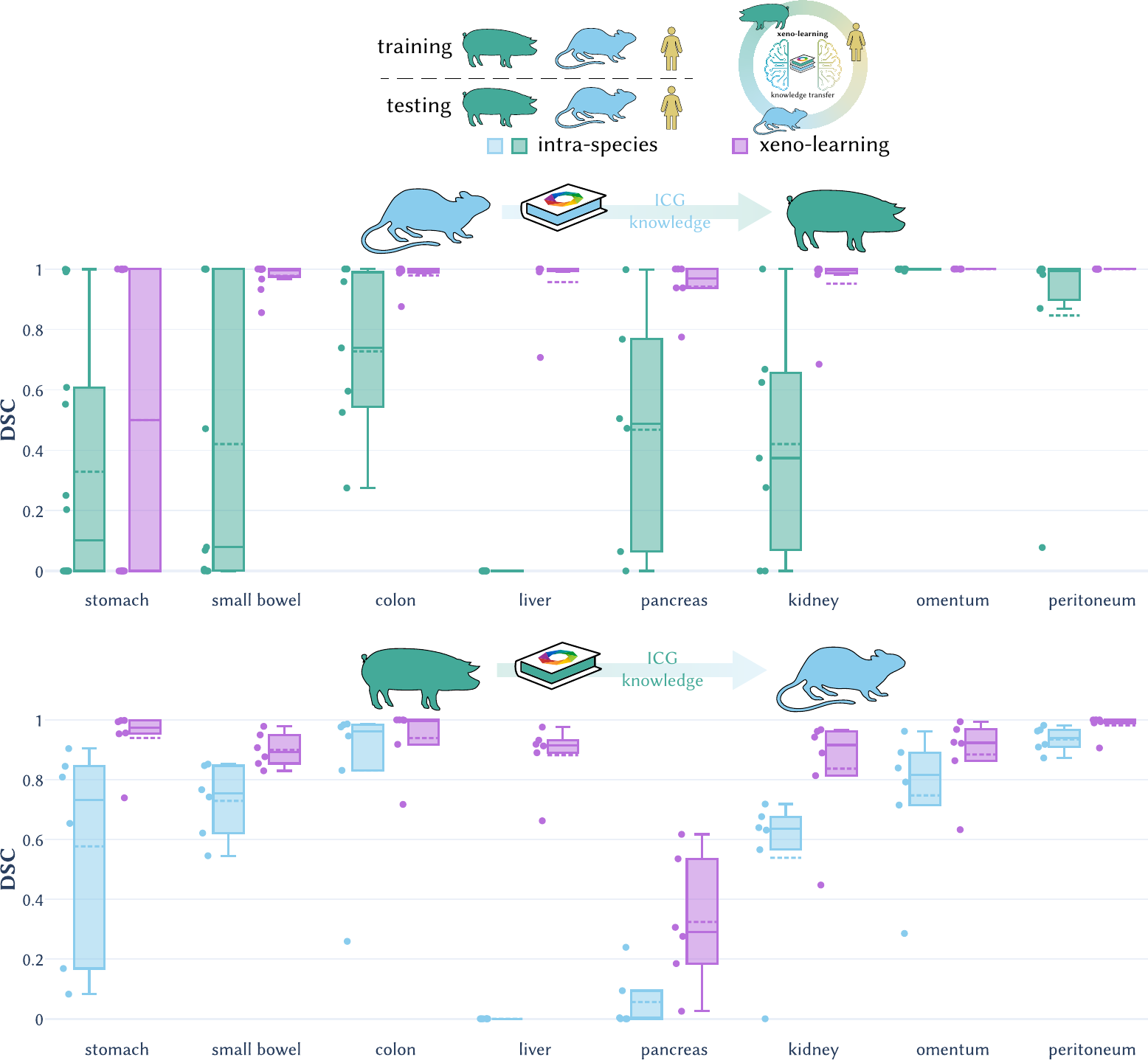}
    \caption{\textbf{Xeno-learning recovers segmentation performance after \acf*{icg} fluorescence injection across organs.} While neural networks trained on data without contrast agent fail to generalize to data with contrast agent (green and blue plots), xeno-learning compensates for this issue (purple plots). The segmentation performance was measured via the \acf*{dsc}. Each boxplot depicts the \acf*{iqr} with the median (solid line) and mean (dotted line). The whiskers extend up to 1.5 times of the \acs*{iqr}. Every point represents the segmentation performance of one subject and organ.}
    \label{fig:icg}
\end{figure*}

As shown in \autoref{fig:icg}, neural networks trained on data without contrast agent fail to generalize to data with contrast agent. This holds true across species and organs. In all cases, our xeno-learning method was able to recover the segmentation performance. For the porcine data, our approach led to an average increase of the \ac{dsc} of \varICGImprovementMeanPig with the \varICGImprovementMaxLabelPig showing the biggest improvement of \varICGImprovementMaxPig. Similarly for the rat data, we obtained an average increase of the \ac{dsc} of \varICGImprovementMeanRat with the \varICGImprovementMaxLabelRat showing the biggest improvement of \varICGImprovementMaxRat.

\section{Discussion}
With this systematic, in-depth analysis of spectral organ differences across species, we are the first to demonstrate that spectral knowledge transfer across species is possible, as exemplified by the task of surgical scene segmentation and the transfer of different perfusion or \ac{icg} states. Based on -- to the best of our knowledge -- the largest \ac{hsi} database containing \varTotalImages images from \varTotalSubjects subjects in three species from which \varTotalImagesSemantic images have been fully semantically annotated with 12 classes, we derived the following key findings:

\begin{enumerate}
    \item \textbf{Inter-species differences of organ spectra are significant}: Different organs across species do not display uniform spectral organ fingerprint features. When considering the species, the subject, and imaging-related factors as sources of spectral variability, the species even represents the main source of variability for certain wavelengths and organs. This heterogeneity causes state-of-the-art segmentation networks to fail when applied across species.
    \item \textbf{Shared mechanisms manifest as comparable relative spectral changes across species}: Substantial \emph{absolute differences} of spectral organ signatures across species render naive approaches of knowledge transfer infeasible. However, our study revealed that relative changes in spectra resulting from pathologies or \ac{icg}-affected organs exhibit similarities across species.
    \item \textbf{Xeno-learning enables knowledge transfer across species}: A proof-of-concept study revealed that the proposed paradigm of learning across species to compensate for quantitative and qualitative data shortage is feasible. From a methodological perspective, the enabling idea was to transfer relative spectral changes via a novel means of \emph{physiological} data augmentation.
\end{enumerate}

\subsection{Strengths and limitations of method}
One of our key observations is that organ differences are high between species and that segmentation networks struggle when applied to another species. Despite that fact, our proposed xeno-learning method is still able to improve the segmentation on malperfused and \ac{icg}-affected spectra. A key to this success lies in the fact that our xeno-learning method is independent of the absolute organ differences between the species as it captures the relative change from a physiological to a malperfused or \ac{icg}-affected organ.

In this proof-of-concept study, we model relative spectral changes resulting from pathologies or surgical manipulation states via a linear transformation which is applied separately to each pixel in the image and which consists only of a manageable number of learnable parameters ($100 \cdot 100 + 100 = \num{10100}$ per image pair). This restriction naturally reduces the risk of overfitting, as there is less freedom to account for unwanted side effects (e.g., from outliers in the data) during the optimization process of the transformation. Though simple by design, it has proven to be a powerful tool to model perfusion changes. Its linearity substantially facilitates computation (affine operation), with only marginal computational overhead during training.
However, these restrictions have the disadvantage that the transformation cannot learn non-linear changes. Furthermore, several applications may require the learning of texture changes and thus require the method to be expanded to larger patches. More specifically, by replacing pixel-wise transformations with patch-wise transformations, it could be possible to alter the texture of tissues using image-to-image translation techniques. However, to keep the computational complexity manageable, it may be necessary to use relatively small models.

Our experiments in \autoref{fig:malperfused} clearly demonstrate that malperfused organs are out of distribution with respect to the physiological training data. While this effect did not lead to a radical performance drop for the specific downstream task of semantic segmentation for all organs, our analyses show that the augmented data captures the malperfused data substantially better.

\subsection{Strengths and limitations of the study}
The fact that spectra differ across species is not surprising as the basis of spectral tissue properties is the biomolecular composition of the organism, which is highly different between species \autocite{10.1016/C2014-0-03145-0}. The most widely studied and spectrally influential protein is hemoglobin which is known to be built differently between species, hence exhibiting different spectral properties under varying environmental conditions and affecting organs differently \autocite{10.1038/s41598-021-04112-y,10.1117/1.JBO.25.9.095002}. Analogous principles apply for other chromophores such as melanin, myoglobin, or bilirubin \autocite{10.1177/039139880102400609}. At the same time, it is also known that there are shared pathophysiological mechanisms across species. This includes the characteristic changes on the absorbance curve of hemoglobin \autocite{10.1117/1.JBO.27.7.074708,10.1364/ol.34.001525,10.1016/j.bcmd.2017.10.006}. In this paper, we showed that such shared pathophysiological mechanisms manifest as comparable relative spectral changes across species, which can then be used for xeno-learning.

While we extensively analyzed the spectral characteristics of 11 physiological organs, we restricted our xeno-learning approach to two exemplary scenarios: correcting the segmentation of kidneys despite different perfusion states (physiological and malperfused) and fluorescent dye application. This decision was made for two main reasons: (1) Kidneys offer a very standardized and controlled setting for inducing pathological organ states through their singular hilar blood supply by clamping or contrast agent application, and (2) these are relevant clinical states regularly encountered during kidney tumor surgery or kidney transplantation. Additionally, kidney pathology has been of interest in other studies on  differentiating arterial ischemia and venous congestion \autocite{10.1038/s41598-024-68280-3}, assessing kidney function in human kidney allotransplantation \autocite{10.1097/SLA.0000000000004429}, or monitoring oxygen saturation during normothermic machine perfusion \autocite{10.5194/jsss-5-313-2016}.

Generally, opportunities for systematic and standardized \ac{hsi} data acquisition are highly limited in humans for ethical and regulatory reasons. Furthermore, human data rarely captures spectral changes in isolation without the influence of other pathologies (e.g., thrombosis), surgical interventions (e.g., cauterization), or side effects from surgical procedures such as kidney transplantations (e.g., medication or host-versus-graft reactions). Organ surfaces may also be less exposed than in animal experiments, such as in cases of adherent renal fat. All these factors make human data acquisition and processing especially challenging. They may also explain the generally lower segmentation performance of neural networks on human data shown in \autoref{fig:domain_shift_performance_dsc} (b).  We addressed the human data acquisition issues with xeno-learning. The animal malperfusion states and \ac{icg} data, induced with high standardization by hilar clamping and systemic intravenous \ac{icg} application, possess a high temporal resolution, as evidenced by the number of image data points in \autoref{fig:perfusion_performance}. Knowledge acquired through these systematic experiments has successfully been transferred to a human application.

\subsection{Our method in the context of domain adaptation}
Our method can be classified as a domain adaptation technique, as it leverages data from both the source and target domains to enhance performance in the target domain \autocite{10.1109/TKDE.2022.3178128}. It is specifically designed for cross-species knowledge transfers in the context of spectral data. We intentionally distinguish between the information we aim to transfer—relative changes in the spectra—while explicitly avoiding the transfer of other types of information, such as absolute spectral reflectance values which are known to be different for each species (cf. \autoref{fig:malperfused}). This is in contrast to domain adaptation methods which directly make use of the absolute spectral values together with the domain information. A comparison of our method with a domain adaptation technique can be found in \autoref{fig:perfusion_performance_domain}.

Unlike other domain adaptation methods, our approach promotes data sharing without requiring access to the source data, as only the learnt relative differences are needed. For example, we provide the necessary projection matrices for applying our physiologically-based data augmentation, enabling anyone to utilize our method on their spectral data.

\subsection{Comparison to other (multi-species) studies}
Previous studies which included multiple species focused purely on spectral analysis without the aim to transfer knowledge in the context of \ac{ai} training. For instance,  \autocite{10.1097/js9.0000000000001849}, \autocite{10.1038/s41598-024-68280-3} and \autocite{10.1515/cdbme-2020-0012} compared spectra from humans and pigs with a focus on the gastric conduit, kidney perfusion states, and on organs of the biliary system, respectively. \autocite{10.1097/js9.0000000000001849} included gastric conduit data from 10 patients and found comparable \ac{sto2} values across species for the malperfusion subgroup. The kidney results of \autocite{10.1038/s41598-024-68280-3} from 17 patients indicate similarities in the malperfusion spectra in line with our results. The presented data partly overlaps our data. While only healthy organs of the biliary system from the seven patients were considered in \autocite{10.1515/cdbme-2020-0012}, significant spectral differences between humans and pigs could be observed, further supporting our findings. It should be noted that these works, however, used significantly lower sample sizes in the range of 7 to 17 patients, while our cohort included \varTotalSubjectsHuman patients. Furthermore, none of them used data from one species for an application to another species.

In the domain of \ac{hsi}, surgical scene segmentation with fully-semantically annotated \ac{hsi} has, until now, been limited to porcine models. The largest study \autocite{10.1007/978-3-031-43996-4_59,10.48550/arxiv.2408.15373} so far comprises 600 images from 33 pigs annotated with 19 classes, with the data partly overlapping our data. The reported \ac{dsc} and \ac{nsd} scores are in agreement with our porcine and rat models (\autoref{fig:domain_shift_performance_dsc} (b) and \autoref{fig:domain_shift_performance_nsd}). Results and solutions regarding the robustness of \ac{hsi} models towards geometrical domain shifts in this study were incorporated into our models. To our knowledge, we were the first to explore the challenge of full semantic scene segmentation with \ac{hsi} in humans.

Human tissue recognition in general has been performed ex-vivo and in-vivo \autocite{10.3390/cancers11060756}. The largest study is from the EX‑MACHYNA trial \autocite{10.1007/s00464-024-10880-1} involving 169 patients undergoing elective open abdominal surgery from two medical centers annotated with 13 tissue classes. While annotations are not fully semantic, they closely follow organ boundaries. The reported \ac{dsc} of 0.80 is in agreement with our reported \ac{dsc} of \varDSCInSpeciesHuman on human data (cf.\ \autoref{fig:domain_shift_performance_dsc} (b)). Here, the main differences to our study lie in the fact that we included malperfused and \ac{icg} data as well as data from three species and explicitly leverage animal data to make the segmentation performance of human data more robust.

Perfusion shifts in human kidney spectra have been analyzed in \autocite{10.1126/sciadv.add6778}. They monitored human patients undergoing partial nephrectomy using a video-rate multispectral imaging device to assess perfused and malperfused kidneys. Their low-dimensional projections, obtained via \ac{pca}, consistently demonstrated that the spectral shifts from physiological to malperfused kidneys followed a uniform direction. These findings are in strong concordance with our results presented in \autoref{fig:malperfused} ©, suggesting that this behavior is consistent across both multispectral and hyperspectral data, as well as across different species.

\subsection{Impact and future directions}
While our xeno-learning paradigm primarily demonstrates the potential for enhanced generalizability across species, it also holds significant promise for the generalization of other variations. For instance, insights into disease models, various pathologies, or imaging conditions observed in one species can potentially be transferred to others. One prominent example of a use case is cancer, a disease marked by differences in perfusion and oxygenation between physiological and pathological tissue. Overall, the expansion of xeno-learning to different use cases will require users to identify shared mechanisms across species and use that prior knowledge to enable the knowledge transfer for \ac{ai}.

Similarly, while our proposed augmentation method is primarily utilized for xeno-learning, its application is by no means limited to cross-species learning. From the perspective of transfer learning in \ac{ai} systems, the method can be broadly employed to identify variations within a source data distribution and subsequently integrate these insights during training within a target data distribution. For instance, observed tissue necrosis, tissue fibrosis, or tissue inflammation in one population can be effectively applied to another population.

Notably, our xeno-learning method employs a simple linear model to transform between two spectral states – in our case, physiological and malperfused or ICG. Due to its simplicity, this method could contribute to the understanding of such spectral changes, for example by analyzing the properties of our transformation matrices. This is especially crucial for human data, which is often confounded by therapy effects, comorbidities, or surgical procedures not present in animal experiments \autocite{10.48550/arxiv.2106.08445,10.48550/arxiv.2408.09873}.

Our results can impact the planning of future animal studies and their potential to be applied to humans. For example, since rat liver is closer to human liver than pig liver, the rat model may be more appropriate for future studies targeting liver pathologies. In contrast, for other organs such as pancreas, neither the pig nor the rat model would be an appropriate candidate.

\subsection{Conclusion}
In summary, we pioneered a new concept of knowledge transfer in the general context of \ac{ai}-based image analysis. Our study with \ac{hsi} data from three species demonstrated the potential of large-scale secondary use of preclinical animal data for humans. The resulting ethical, monetary, and performance benefits of the proposed knowledge transfer paradigm promise a high impact of the methodology on future developments in the field. For maximum impact, our code and pretrained models will be made publicly available.

\section{Methods}
\subsection{Data Collection}
The \ac{hsi} animal data was collected at Heidelberg University Hospital following approval from the Committee on Animal Experimentation of the regional council Baden-Württemberg in Karlsruhe, Germany (G-161/18, G-262/19 and G-62/23). The animals were treated in accordance with German laws for animal use and care, as well as the directives of the European Community Council (2010/63/EU).

The \ac{hsi} human data was obtained during the SPACE trial (SPectrAl Characterization of organs and tissuEs during surgery) at Heidelberg University Hospital, with approval from the Ethics Committee of the Medical Faculty of Heidelberg University, Germany (S-459/2020). The trial adhered to the ethical principles of the Declaration of Helsinki \autocite{ethics_Helsinki_2003} and the principles of Good Clinical Practice \autocite{good_clinical_practice_ich2001}. The trial’s reporting followed the recommendations of the Consolidated Standards of Reporting Trials (CONSORT) guideline \autocite{randomised_trials_moher2010}. The SPACE trial was registered with the Research Registry (researchregistry6281) on November 23, 2020.

\subsection{Hyperspectral image acquisition}
The \ac{hsi} camera system Tivita\textsuperscript{\textregistered} Tissue (Diaspective Vision GmbH, Am Salzhaff, Germany) was utilized to collect the \ac{hsi} data. This system captured hyperspectral images in a push-broom manner with a spectral resolution of approximately \SI{5}{\nm}, covering the spectral range from \SI{500}{\nm} to \SI{1000}{\nm}. The resulting data cubes have dimensions of $640 \times 480 \times 100$ (width $\times$ height $\times$ number of spectral channels). The camera system imaged an area of about $20 \times \SI{30}{\cm}$. An integrated distance calibration system, consisting of two light marks that overlap when the distance is correct, ensured an imaging distance of around \SI{50}{\cm}. The image acquisition process took approximately seven seconds.

In addition to the \ac{hsi} data cubes, the camera system computed functional parameters such as \ac{sto2} \autocite{10.1515/bmt-2017-0155}. Furthermore, RGB images were reconstructed from the \ac{hsi} data by combining spectral channels that capture red, green, and blue light. More technical details on the hardware and the performed calculations can be found in \autocite{10.1515/bmt-2017-0145}.

To prevent spectral distortion from stray light, all other light sources were turned off during image capture, and window blinds were closed. Motion artifacts were minimized by (1) mounting the camera on a swivel arm to keep it stationary during image capture, thus eliminating camera motion, and (2) capturing images from static scenes with no surgeon-induced object movements. Consequently, any motion artifacts would only be due to natural causes such as respiration and heartbeat. The camera perspectives were chosen to provide a clear view of all organs of interest in the scene.

For the pig and rat species, standardized recordings were carried out with a predefined image acquisition protocol which is described in \autocite{10.1038/s41597-023-02315-8}. This data is used in \autoref{fig:mixed_effect_analysis}.

For the acquisition of animal data, the organs were mobilized and prepared to enable the spectral recording of representative tissue surfaces. Time series data was acquired by taking images every 30 to 60 seconds after the clamping procedure.

For the purpose of this work, the following types of visceral organ data were recorded:physiological (all three species),  malperfused (all three species) and with contrast agent \ac{icg} (mice and porcine models): Note that there was no opportunity to record \ac{icg} data in human patients for ethical reasons.

The animal malperfused organ data were acquired by vascular clamping of the infradiaphragmal aorta and caval vein in order to induce visceral malperfusion in a highly controlled and standardized manner. Hemodynamic stabilization and tissue oxygen desaturation over two minutes was awaited before measurements.
 
The human data was acquired in a clinical trial during which malperfused organs were recorded whenever they occurred intraoperatively, such as during kidney transplantation or acute organ ischemia due to embolic or thrombotic events or iatrogenic preparation and resection. Human malperfused data was therefore available for colon, liver, kidney and spleen. Consequently, analysis of malperfused data in the animal species was also restricted to this organ selection.

The animal \ac{icg} data was acquired by body-weight adapted intravenous application of commercial \ac{icg}. Pigs received 25 mg of \ac{icg} injected into peripheral vein catheters, while rats received 2.5 mg directly injected into the caval vein. Pharmacological invasion and systematic distribution over one minute was awaited before measurements.

\subsection{Hyperspectral image annotation}
From the \varTotalImages images, \varTotalImagesSemantic were fully-semantically annotated. For the remaining \varTotalImagesPolygon images, polygon annotations were performed to annotate highly representative areas of the organ. All annotations were based on the reconstructed RGB image.

Polygon annotations were performed in the same manner as described in \autocite{10.1038/s41598-022-15040-w} and the standardized recordings for the pig species are publicly available \autocite{10.1038/s41597-023-02315-8}. Polygon annotations were used in the mixed effects analysis of \autoref{fig:mixed_effect_analysis} and for several malperfused tissue annotations.

The semantic physiological porcine annotations are the same as described in \autocite{10.1016/j.media.2022.102488} where the semantic annotation process was performed by two different medical experts. Conflicts were revised by the same two medical experts. For all remaining semantic annotations, the annotation process was conducted by a team of medical experts using the \ac{mitk} \autocite{MITK_Team_MITK_2024}. To ensure consistent labeling, all annotations were reviewed by the same medical experts. The segmentation networks were only trained on images with semantic annotations.

From the annotations, only the classes stomach, small bowel, colon, liver, pancreas, kidney, spleen, omentum, lung, skin, peritoneum, and background were selected for this study, matching the organs which are available for all three species (see below). All remaining classes were not included in the analysis.

\subsection{Data preprocessing}
To mitigate sensor noise and transition the acquired \ac{hsi} data from radiance to reflectance, the raw \ac{hsi} data cubes were automatically calibrated using pre-recorded white and dark calibration files by the camera system, as outlined in \autocite{10.1515/bmt-2017-0155}. Calibration of the camera was performed before each surgery by taking a new white and dark image to compensate for various sources of signal distortion, such as attenuation effects of the light source \autocite{10.1016/b978-0-444-63977-6.00021-3}.

After exporting the \ac{hsi} cubes from the camera system, each pixel in the \ac{hsi} cube was L1-normalized across the spectral channels to account for multiplicative illumination changes, such as those caused by fluctuations in the measurement distance.

\subsection{Data statistics}
The hyperspectral imaging database used in this study is composed of \varTotalImages images from \varTotalSubjects subjects in three species. Annotation has been performed for 12 classes (11 organs and background). These organs were selected as they were available for all three species due to anatomical, anesthesiological and technical considerations. Malperfused tissues were captured on \varTotalImagesMalPig pig images, \varTotalImagesMalRat rat images, and \varTotalImagesMalHuman human images. Tissues following \ac{icg} injection were captured on \varTotalImagesICGPig pig images and \varTotalImagesICGRat rat images. The remaining \varTotalImagesPhysPig pig images, \varTotalImagesPhysRat rat images and \varTotalImagesPhysHuman human images show physiological tissues. A detailed overview of the database is given in \autoref{fig:data_statistics}.

\subsection{Segmentation models}
The employed segmentation networks are the same as described in \autocite{10.1016/j.media.2022.102488} with the organ transplantation extension proposed in \autocite{10.1007/978-3-031-43996-4_59,10.48550/arxiv.2408.15373}. In short, the full \ac{hsi} data cube is passed on to a U-Net with an efficientnet-b5 encoder pre-trained on the ImageNet dataset. Dice loss and \ac{ce} loss are equally weighted and computed for all valid pixels inside a batch, i.e., every pixel which does not belong to one of the ignored classes. 

The same hyperparameters were used for all models. Adam \autocite{adam_kingma2015} was used as an optimization algorithm with an exponential learning rate scheme (initial learning rate: $\eta=0.001$, decay rate $\gamma=0.99$, Adam decay rates $\beta_1=0.9$and $\beta_2=0.999$). Training was carried out with a batch size of 8 for 100 epochs with each epoch consisting of 500 images. During the last 10 epochs, stochastic weight averaging was applied \autocite{swa_izmailov2018}. Underrepresented classes were oversampled to ensure an equal class distribution.

During training, images were augmented to increase the size and the diversity of the training data. First, the same affine transformations as described in \autocite{10.1016/j.media.2022.102488} (shift, scale, rotate and flip operations) were applied. Then, target tissues in the images (kidney in our study) were transformed with our proposed xeno-learning method. Finally, the organ transplantation method proposed in \autocite{10.1007/978-3-031-43996-4_59,10.48550/arxiv.2408.15373} followed by a L1 re-normalization of the image was applied.

\subsection{Training and validation setup}
A similar training and validation setup was used for all networks. Validation was carried out based on a nested cross-validation scheme to provide a more robust performance estimation based on the entire dataset \autocite{10.1186/1471-2105-7-91}. The number of outer folds was set to 3 and the number of inner folds was set to 5. The folds were generated based on iterative stratification for multi-labeled data \autocite{10.1007/978-3-642-23808-6_10} to ensure a similar label distribution across folds. Final predictions for an image were obtained by ensembling the softmax output from all available networks.

To prevent model overfitting in the standard machine learning paradigm, it is crucial to evaluate methods using an untouched test set. We extend this principle by incorporating completely unseen \textit{tasks and species} in our evaluation framework. During the development of our method, we focused exclusively on one task (malperfusion) and data from only the pig species, phrasing the problem as a pig2pig malperfusion task, where the domain shift stems from unseen individuals. After finalizing our method, we first transferred the approach to further species (rats and humans) and then to another knowledge transfer task. Crucially, pig \ac{icg} data, as well as all rat and human data, were not used at any stage during the development of our method.

Following the recommendations in \autocite{10.1038/s41592-023-02151-z} and to overcome the limitations of individual metrics, we assessed the segmentation performance via the overlap-based \ac{dsc} and the boundary-based \ac{nsd}. The class-specific thresholds for the \ac{nsd} were set to the same values as reported in \autocite{10.1016/j.media.2022.102488}. Metric scores were always computed per class and then hierarchically aggregated towards a final class-level score. That is, class-scores were averaged first across all images of the same subject and then across all subject-level class scores.

For the results of \autoref{fig:perfusion_performance}, only the \ac{dsc} is reported because some of the annotations of malperfused organs were performed via polygon annotations (cf.\ annotations section). The goal of these polygon annotations was to annotate representative areas of an organ but not to strictly follow the organ boundaries. Hence, the computation of the \ac{nsd} (which assesses boundary-agreement) was not reasonable.

Confidence intervals reported in this study reflect the subject-level sampling variability of the data. After the aggregation towards subject-level scores, bootstrapped sampling was performed with replacement 1000 times for each organ and the sampled subject-level scores were averaged to retrieve an organ-level score per bootstrap sample. After aggregating (e.g., mean or median) across organs for each bootstrap sample, \SI{95}{\percent} confidence intervals were calculated.

\subsection{Linear mixed model analysis}
Separate linear mixed models for each wavelength and organ were employed to analyze explained variation in order to evaluate the relevance of factors contributing to changes in the observed spectrum (\autoref{fig:mixed_effect_analysis}). The proportion of explained variance was derived through the empirical decomposition of explained variation based on the variance components version of the mixed model \autocite{10.1101/2019.12.28.890061}.

More precisely, linear mixed models were fitted for each organ and wavelength separately, with fixed effects for the factor angle and the factor species as well as random effects for the factor subject and the factor image:
\begin{equation}
\begin{aligned}
    \text{reflectance}_{ijk} = \alpha &+ \text{species}^\top_{ijk} \cdot \beta \\
    &+ \text{angle}^\top_{ijk} \cdot \theta \\
    &+ \delta_i + \gamma_{ij} + \varepsilon_{ijk},
\end{aligned}
\end{equation}
for repetition $k = 1, \ldots, 3$ of image $j = 1, \ldots, n_i$ of animal $i=1, \ldots, 24$ ($11$ pigs and $13$ rats). 

The number of images $n_i$ varied per animal and organ. Here, $\alpha$ denotes a fixed intercept, $\text{species}^\top_{ijk}$ is a row vector of length $2$ indicating the species rat or pig with $\beta$ denoting the corresponding fixed effect. Similarly, $\theta$ is a vector of fixed effects corresponding to the camera angles (\enquote{perpendicular to tissue surface}, \enquote{25 degree from one side}, \enquote{25 degree from the opposite side}). The random intercept $\delta_i\sim \mathcal{N}(0, \sigma_\delta^2)$ describes animal specific variation, and the random intercept $\gamma_{ij}\sim \mathcal{N}(0, \sigma^2_\gamma)$ describes image specific variations. The residuals $\varepsilon_{ijk}\sim \mathcal{N}(0, \sigma_\varepsilon^2)$ capture the variability between repeated recordings of the same image. Within the model, we assumed that the random effects and the residuals are stochastically independent. 

Additionally, \SI{95}{\percent} pointwise confidence intervals were obtained based on parametric bootstrapping with 500 replications for an indication of the uncertainty in the relevance estimates.

\subsection{Xeno-learning}
The driving insight of our concept is that shared  (e.g. pathophysiological) mechanisms manifest in comparable relative spectral changes. In our approach (cf.\ \autoref{fig:method_concept}), we leverage this insight to transfer knowledge across species through data augmentation. Specifically, we first learn the effect of certain interventions, such as clamping or contrast agent injection, in the source species and then apply this knowledge in the target species. 

\paragraph{Learning relative changes in the source species}
We encode relative spectral changes in a transformation matrix that can be applied to any species. For the specific case of perfusion, a set of linear transformations $t_i(s_p)$ are learnt that transform physiological spectra $s_p$ to malperfused spectra $s_m$ (the transformation is applied independently for each spectra). To cover a variety of different perfusion states, we learn a whole set of transformations each of which represents the spectral change between physiological and malperfused kidneys. Each transformation is a linear model represented by two parameters: a weight matrix $W_i \in \mathbb{R}^{100\times100}$ and a bias vector $b_i \in \mathbb{R}^{100}$, in analogy of a multivariate linear regression model, so that the transformation is defined as
\begin{equation}
	\label{eq:spectra_transformation}
	t_i(s_p) = W_i \cdot s_p + b_i
\end{equation}
To learn the transformation $t_i$, image pairs $(p_i, m_i)$ are randomly selected in the source species, consisting of an image $p_i$ showing a physiological kidney and an image $m_i$ showing a malperfused kidney. The images do not necessarily need to come from the same subject (the transformation will always be applied to different subjects in the target species).

The transformation parameters $W_i$ and $b_i$ should be based on the spectra in the image pair $(p_i, m_i)$. However, since the number of spectra for a specific organ in each image is usually different (and hence no match between the spectra $s_p$ and $s_m$ is available), $W_i$ and $b_i$ cannot be optimized directly in closed form. Instead, the following indirect optimization scheme is applied for each selected image pair $(p_i, m_i)$:
\begin{enumerate}
    \item The parameter weight matrix is initialized with the identity matrix $W_i = I$ and the bias with zeros $b_i=0$. Hence, the initial transformation $t_i(s_p)$ does not modify the spectra.
    \item The MSE loss is used to compare the malperfused spectra $s_m, \in \mathcal{S}_m$ (set of all spectra in the malperfused image $m_i$) with the physiologically-transformed spectra $t_i(s_p)$ with $s_p \in \mathcal{S}_p$ (set of all spectra in the physiological image $p_i$). The loss consists of three components: (1) two histograms each with 50 bins: $h_m$ computed from the normalized reflectance values from all $s_m$ and $h_t$ computed from the normalized reflectance values from all $t_i(s_p)$, (2) the mean spectrum $\hat{s}_m$ and $\hat{t}_i(s_p)$ and (3) the standard deviation spectrum $\tilde{s}_m$ and $\tilde{t}_i(s_p)$.
    \item The linear model is fitted iteratively in 100 steps with the Adam optimizer (learning rate: $\eta=0.001$, Adam decay rates $\beta_1=0.9$ and $\beta_2=0.999$) while gradually adapting the parameters $W_i$ and $b_i$.
\end{enumerate}

\paragraph{Transferring knowledge}
After the optimization phase in the source species, all learnt changes, encoded in the parameters $W_i$ and $b_i$, can be applied in the target species. For increased data variety, this is done dynamically during the training process as augmentation. For every image in the batch, the physiological kidney spectra $\mathcal{S}_p$ are selected using the available segmentation mask. Then, a parameter set $j$ is randomly selected and the corresponding parameters $W_{j}$ and $b_{j}$ are used to transform every kidney spectrum $s_p \in \mathcal{S}_p$ via \autoref{eq:spectra_transformation}. In order to cover many different perfusion states, the transformed spectra are linearly interpolated with a randomly selected weight $\lambda \in [0;1]$
\begin{equation}
	s = \left( 1 - \lambda \right) \cdot s_p + \lambda \cdot t_{j}(s_p)
\end{equation}
$s$ denotes the replaced spectra which are shown to the network. In total, this augmentation is applied to an image with probability $p=0.8$ so that some unaltered kidney spectra are also shown during the training process.

\subsection{Distribution comparison}
The extended training distribution shown in \autoref{fig:perfusion_projections} was derived by a training process comprising 100 epochs with 500 images per epoch, during which our data augmentation was applied in the same manner as during the training of the segmentation networks. For each organ in each image seen during the training phase, the median spectrum was computed before and after augmentation. The resulting distributions of median spectra are referred to as baseline training distribution (black distribution in \autoref{fig:perfusion_projections}) and extended training distribution (purple distribution in \autoref{fig:perfusion_projections}).

To compare the baseline training distribution with the extended training distribution, two-dimensional projections of all median spectra were computed using \ac{pca}, where the principal components were derived from the baseline training distribution. The median spectra from the extended training distribution were subsequently projected into this same feature space. Both distributions were modeled using kernel density estimation. Distances were calculated by measuring the Euclidean distance between each median spectrum in the test set and the nearest median spectrum in either the baseline or extended training distribution. These distances were then hierarchically averaged, first across all images belonging to a single subject, and subsequently across all subjects.

\subsection{Reporting summary}
Further information on research design is available in the Nature Research Reporting Summary linked to this article.

\backmatter


\section*{Declarations}
\paragraph{Acknowledgements}
This project has received funding from the European Research Council (ERC) under the European Union’s Horizon 2020 research and innovation programme (project NEURAL SPICING grant agreement No. 101002198) and the National Center for Tumor Diseases (NCT) Heidelberg's Surgical Oncology Program. It was further supported by the German Cancer Research Center (DKFZ) and the Helmholtz Association under the joint research school HIDSS4Health (Helmholtz Information and Data Science School for Health). This publication was supported through state funds approved by the State Parliament of Baden-Württemberg for the Innovation Campus Health + Life Science alliance Heidelberg Mannheim.

We would like to thank Hannah Gottlieb, Oray Kalayci, Hussein Bahaaeldin, Polina Borisova, Pit Beckius, Lotta Biehl, Fatmanur Yilmaz, Patrick Unverdorben and Laura Mehlan for annotating the data.

\paragraph{Author contributions}
ASF, ABQ, JS, LMH and SS conceived and designed the study. LMH invented the paradigm of xeno-learning and kickstarted the project with FN. ASF, BÖ, CMH, GS, JB, JH, KFK, MD, MSM and SK designed and performed the data acquisition. ASF, ABQ, JB, JH, JS and SS supervised the annotation of the data. JS developed the data augmentation method and implemented the experiments. ASF, ABQ, JS, LMH and SS analyzed the data. AK, MW and NS provided the linear mixed model analysis. ASF, JS, LMH and MT wrote the manuscript. ASF, ABQ, JS, LMH, MD, MT, NS and SS revised the manuscript. All authors have read and approved the final manuscript.

\paragraph{Data availability}
The animal data and pretrained models for the main results of this study will be made publicly available and are accessible through our code repository at \href{https://github.com/IMSY-DKFZ/htc}{https://github.com/IMSY-DKFZ/htc}.

\paragraph{Code availability}
All code will be made publicly available in our GitHub repository at \href{https://github.com/IMSY-DKFZ/htc}{https://github.com/IMSY-DKFZ/htc}, including pretrained models.

\paragraph{Competing interests}
Lena Maier-Hein worked with the medical device manufacturer KARL STORZ SE \& Co. KG in the projects \enquote{InnOPlan} and \enquote{OP 4.1}, funded by the German Federal Ministry of Economic Affairs and Energy (grant agreement No. BMWI 01MD15002E and BMWI 01MT17001E) and \enquote{Surgomics}, funded by the German Federal Ministry of Health (grant agreement No. BMG 2520DAT82D).

\printacronyms

\bibliography{sn-bibliography}

\begin{appendices}
\section{Extended data}

\begin{figure*}[hp]
    \centering
    \includegraphics[width=\linewidth]{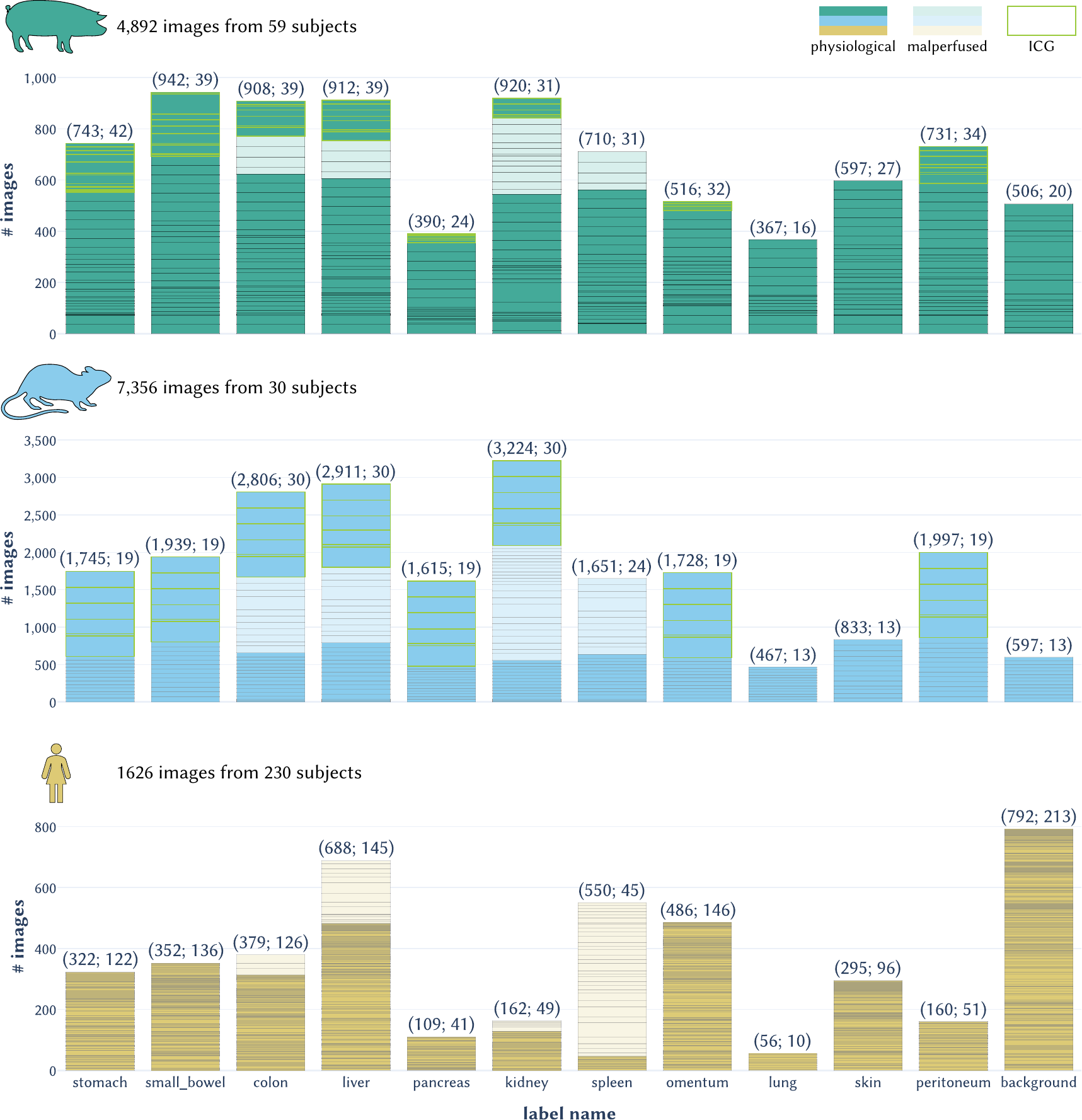}
    \caption{\textbf{Overview of our multi-species dataset.} For each species, the number of physiological, malperfused (light color) and \acf*{icg} (green border) images are shown. Different subjects are separated via lines. The numbers on top of the bars denote the number of images and subjects for the respective label.}
    \label{fig:data_statistics}
\end{figure*}

\begin{figure*}[hp]
    \centering
    \includegraphics[width=\linewidth]{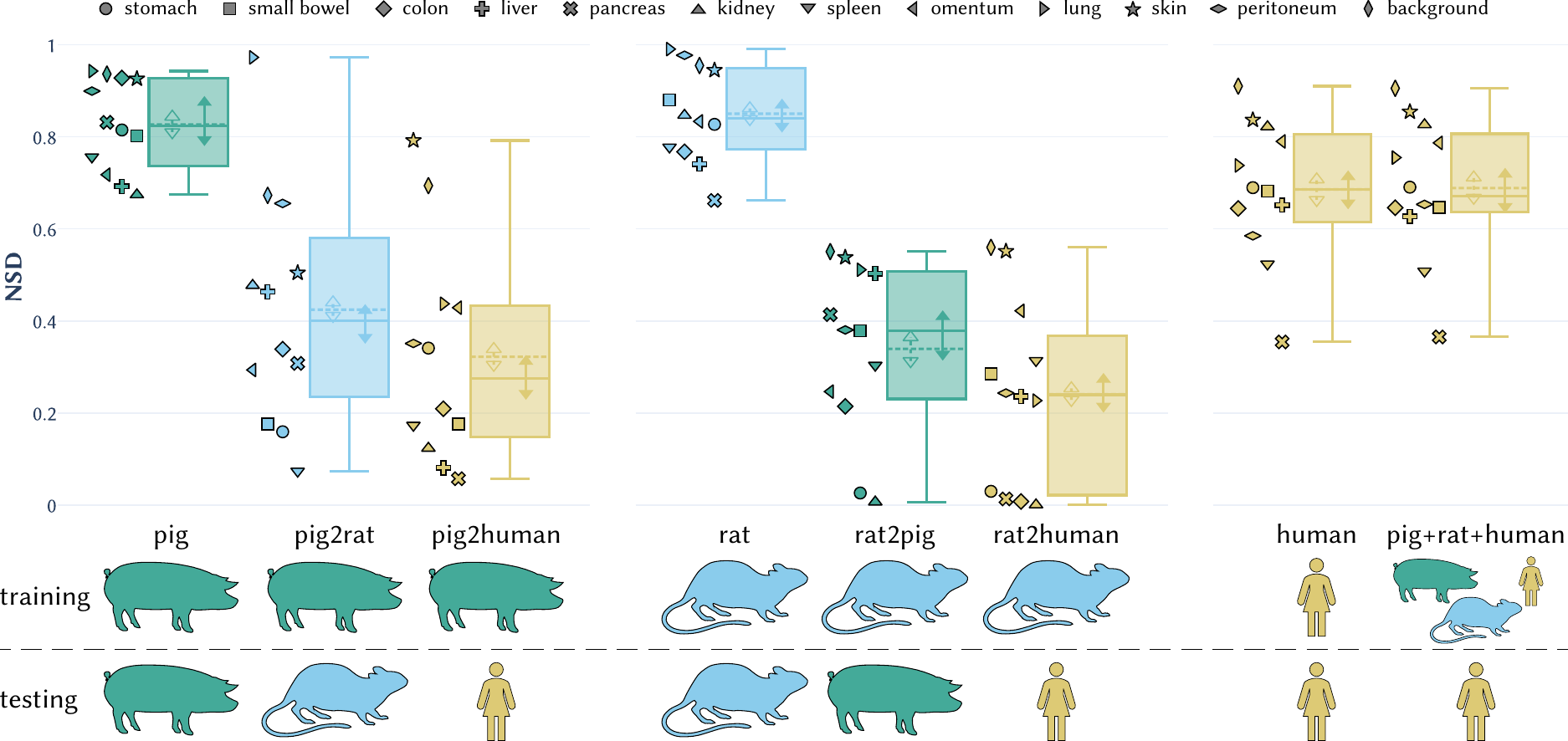}
    \caption{\textbf{Different species exhibit different spectral organ fingerprints, resulting in organ segmentation networks trained on one species failing to generalize towards unseen species.} For each species, the intra-species segmentation performance always surpasses the out-of-species performance. The distributions of hierarchically aggregated class-level \acf*{nsd} scores are shown. Each boxplot depicts the \acf*{iqr} with the median (solid line) and mean (dotted line). Arrows indicate the \SI{95}{\percent} confidence interval of the median and mean based on bootstrapped subject-sampling. The whiskers extend up to 1.5 times of the \acs*{iqr}. Results for the \acf*{dsc} are presented in \autoref{fig:domain_shift_performance_dsc}.}
    \label{fig:domain_shift_performance_nsd}
\end{figure*}

\begin{figure*}[hp]
    \centering
    \includegraphics[width=\linewidth]{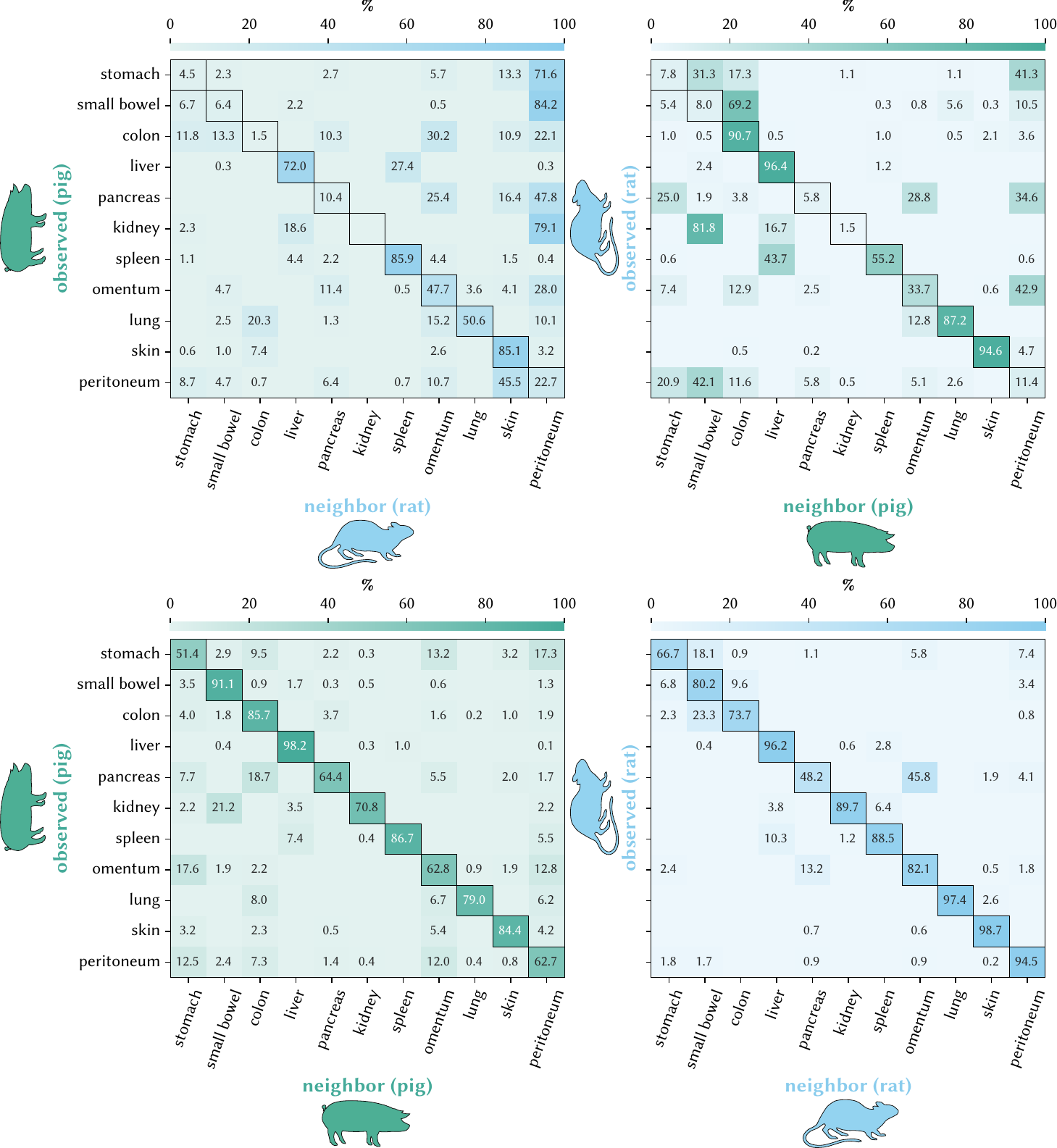}
    \caption{\textbf{Spectra of the same tissue type are not nearest neighbors across species.} Comparison between pig and rat (top left), rat and pig (top right) as well as pig to pig subjects (bottom left) and rat to rat subjects (bottom right) median spectra per organ. For every median spectrum of the observed species, we determined the nearest neighbor in the neighbor species and compared the class labels. The nearest neighbor spectrum is always from a different subject. Each matrix is row-normalized, highlighting how the median spectra from the observed species class are distributed across the nearest neighbor species classes. Human comparisons can be found in \autoref{fig:nearest_neighbor}.}
    \label{fig:nearest_neighbor_suppl}
\end{figure*}

\begin{figure*}[ht]
    \centering
    \includegraphics[width=\linewidth]{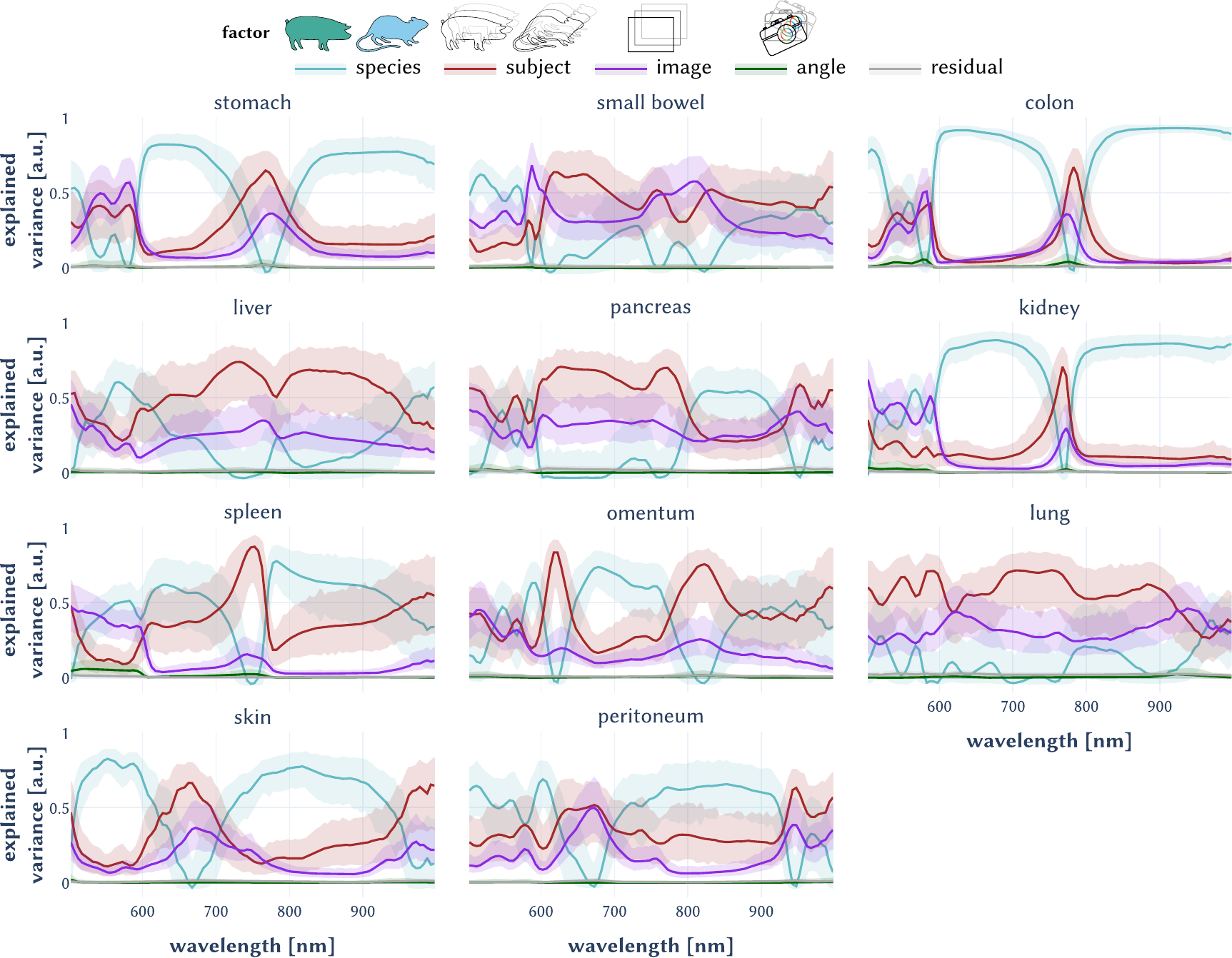}
    \caption{\textbf{Mixed effect analysis on standardized animal data reveals no dominant factor regarding variability across organs.} For each organ, the proportion of variability in reflectance explained by the (i) species, (ii) subject, (iii) image and (iv) angle factors is shown by fitting a linear mixed model to each wavelength and organ independently. The factors describe the relevance of variability resulting from differences in (i) species, (ii) individuals of the same species, (iii) organ positions, (iv) and angles between the organ surface and the camera optical axis, respectively. The residual factor captures variability not explained by the other factors. Shaded areas denote \SI{95}{\percent} confidence intervals based on parametric bootstrap sampling. For some organs and spectral regions, the species factor is most relevant whereas for others the subject factor explains most of the variability.}
    \label{fig:mixed_effect_analysis}
\end{figure*}

\begin{figure*}[hp]
    \centering
    \includegraphics[width=\linewidth]{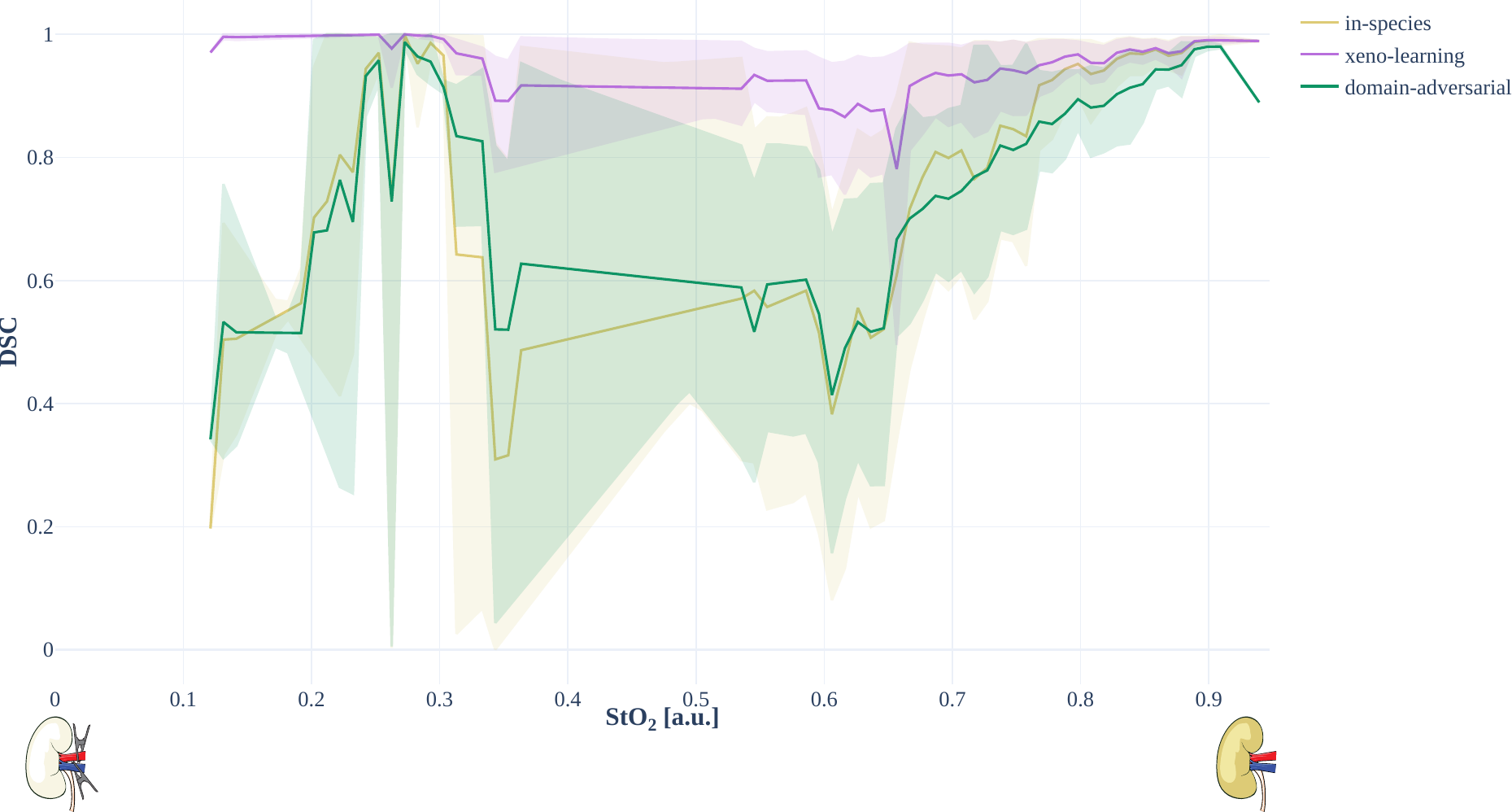}
    \caption{\textbf{Xeno-learning outperforms adversarial domain adaptation}. Similarly to \autoref{fig:perfusion_performance}, the segmentation performance (measured via the \acf*{dsc}) of our xeno-learning approach is compared with an adversarial domain adaptation technique \autocite{JMLR:v17:15-239}. For each subject in the target species, a line was interpolated over all $(\acs*{sto2}, \acs*{dsc})$ values and hierarchically aggregated across subjects. The shaded area denotes the \SI{95}{\percent} confidence interval based on bootstrapped subject-sampling.}
    \label{fig:perfusion_performance_domain}
\end{figure*}

\end{appendices}



\end{document}